\documentclass{article} 
\usepackage{iclr2023_workshop,times}
\pdfoutput=1
\usepackage[frozencache,cachedir=.]{minted}  

\usepackage{amsmath,amsfonts,bm}

\def\eqref#1{equation~\ref{#1}}

\def\1{\bm{1}}

\DeclareMathAlphabet{\mathsfit}{\encodingdefault}{\sfdefault}{m}{sl}
\SetMathAlphabet{\mathsfit}{bold}{\encodingdefault}{\sfdefault}{bx}{n}

\usepackage{amsmath,amsthm,amssymb,amsfonts}
\usepackage{booktabs,array,tabularx}

\usepackage{physics}
\usepackage{multirow}
\usepackage{esvect}
\usepackage{ulem}
\usepackage{cancel}

\usepackage{dsfont}
\usepackage{subcaption}
\usepackage{bm}
\usepackage{svg}

\usepackage[skins]{tcolorbox}
\usepackage{tikz}
\usetikzlibrary{
    shapes, 
    arrows.meta, 
    positioning}

\usepackage{pgfplots}
\pgfplotsset{width=10cm,compat=1.9}

\usepackage{algorithm}
\usepackage{paralist}
\usepackage{minted}

\usepackage{titlecaps}

\newtheorem*{theorem}{Theorem}
\newtheorem{lemma}{Lemma}

\newcommand{\fig}[2]{\includegraphics[width=#2\textwidth]{imgs/#1}}
\newcommand{\parbasic}[1]{\textbf{#1} \hspace{0.5mm}}

\newcommand{\bh}[1]{\mathbf{\hat{#1}}}
\newcommand{\bt}[1]{\mathbf{\tilde{#1}}}

\newcommand{\graphtype}{3D graph}
\newcommand{\receptor}{base graph}
\newcommand{\ligand}{complement graph}
\newcommand{\complex}{complex}
\newcommand{\conditionalEGNN}{Conditional EGNN} 
\newcommand{\conditionalFlow}{Conditional Flow} 
\newcommand{\ccmapping}{Complex-to-Complement Mapping} 
                                                       
\newcommand{\ligsym}{G} 
\newcommand{\recsym}{\hat{G}} 

\usepackage[T1]{fontenc}    
\usepackage{hyperref}
\usepackage{url}

\title{Semi-Equivariant Conditional Normalizing Flows}

\author{Eyal Rozenberg and Daniel Freedman \\
Verily Research\\
Haifa, Israel \\
\texttt{\{eyalrozenberg,danielfreedman\}@verily.com}
}

\iclrfinalcopy 
\begin{document}

\maketitle

\begin{abstract}
We study the problem of learning conditional distributions of the form $p(\ligsym | \recsym)$, where $\ligsym$ and $\recsym$ are two \graphtype{}s, using continuous normalizing flows.  We derive a semi-equivariance condition on the flow which ensures that conditional invariance to rigid motions holds.  We demonstrate the effectiveness of the technique in the molecular setting of receptor-aware ligand generation.
\end{abstract}

\section{Introduction}
Data consisting of sets of three-dimensional points appear in a number of scientific and engineering settings; examples include molecular chemistry, high energy physics, and computer vision.  In many cases, these point sets can be effectively described by a \graphtype{}, that is a graph whose vertices correspond to points in $\mathbb{R}^3$.  In this work, we are concerned with learning conditional distributions of the form $p(\ligsym | \recsym)$, where $\ligsym$ and $\recsym$ are two \graphtype{}s.  Construction of such distributions is useful in various scenarios: in the molecular setting, $\recsym$ will represent a receptor / target and $\ligsym$ will represent a ligand; in the setting of shape completion in computer vision, $\recsym$ will represent the part of the point cloud that we have been given, and $\ligsym$ will represent the completion of this point cloud.

We present a method for learning such conditional distributions based on continuous normalizing flows.  A critical aspect of the problem is to learn a distribution which respects the correct invariances.  Specifically, there must be an invariance to rigid motions which is a kind of ``conditional invariance'' which is expressed jointly in terms of $\ligsym$ and $\recsym$.  Our main theoretical result is that this conditional invariance can be satisfied by a novel form of \textit{semi-equivariance} of the normalizing flow.  We then empirically demonstrate the utility of our technique in the molecular setting: our method allows for effective receptor-aware ligand generation. We discuss related work in Appendix \ref{sec:related_work}.

\section{Semi-Equivariant Conditional Normalizing Flows}
\label{sec:methods}
\subsection{Notation and Goal}
\label{sec:notation}

\parbasic{Notation} A \graphtype{} is given by $\ligsym = (N, V, E, A)$, where $N$ is the number of vertices; $V$ is the list of vertices; $E$ is the list of edges; and $A$ is the set of global graph properties, i.e. properties which apply to the entire graph.  The vertex list\footnote{We use lists, rather than sets, so as to make the action of permutations clear.} is $V = \left( \mathbf{v}_i \right)_{i=1}^N$ where each vertex is specified by a vector $\mathbf{v}_i = (\mathbf{x}_i, \mathbf{h}_i)$; in which $\mathbf{x}_i \in \mathbb{R}^3$ is the position of the vertex, and $\mathbf{h}_i \in \mathbb{R}^{d_h}$ contains the properties of the vertex.  The edges in the graph $\ligsym$ are undirected, and the edge list $E$ is specified by a neighbourhood relationship.  Specifically, if $\eta_i$ is the set of vertex $i$'s neighbours, then we write $E = \left( \mathbf{e}_{ij} \right)_{i<j: j \in \eta_i}$.  The vector $\mathbf{e}_{ij} \in \mathbb{R}^{d_e}$ contains the properties of the edge connecting vertices $i$ and $j$.  Finally, the graph properties are given by $K$ individual properties, i.e. $A = (\mathbf{a}_1, \dots, \mathbf{a}_K)$. 

\parbasic{Rigid Transformations}  The action of a rigid transformation $T \in E(3)$ on a graph $\ligsym$ is given by $T\ligsym = (TN, TV, TE, TA)$ where $TV = \left( T\mathbf{v}_i \right)_{i=1}^N$ with $T\mathbf{v}_i = (T\mathbf{x}_i, \mathbf{h}_i)$.  The other variables are unaffected by $T$; that is $TN = N$, $TE = E$, and $TA = A$.

\parbasic{Permutations} The action of a permutation $\pi \in \mathbb{S}_N$ on a graph $\ligsym$ with $N(\ligsym) = N$ is given by $\pi \ligsym = (\pi N, \pi V, \pi E, \pi A)$ where $\pi V = \left( \mathbf{v}_{\pi_i} \right)_{i=1}^N$ and $\pi E = \left( \mathbf{e}_{\pi_i\pi_j} \right)_{i<j: j \in \eta_i}$.  The other variables are unaffected by $\pi$; that is, $\pi N = N$ and $\pi A = A$.

\parbasic{Goal} We assume that we have both a \receptor{} and a \ligand{}, each of which is specified by a \graphtype{}.  We denote the \receptor{} as $\recsym = (\hat{N}, \hat{V}, \hat{E}, \hat{A})$ and the \ligand{} as $\ligsym = (N, V, E, A)$.  Our goal is to learn a conditional generative model: given the \receptor{}, we would like to generate possible \ligand{}s.  Formally, we want to learn $p(\ligsym | \recsym)$. We want our generative model to respect two types of symmetries, expressed mathematically as:
\begin{equation}
    p(T\ligsym | T\recsym) = p(\ligsym | \recsym) \,\, \forall T \in E(3) 
    \quad \text{and} \quad
    p(\pi \ligsym | \hat{\pi} \recsym) = p(\ligsym | \recsym) \,\, \forall \pi \in \mathbb{S}_{N}, \, \hat{\pi} \in \mathbb{S}_{\hat{N}}
    \label{eq:invariance}
\end{equation}
The first condition says that if we transform both the \ligand{} and the \receptor{} with the same rigid transformation, the probability should not change.  The second condition says that permuting the order of either the \ligand{} or the \receptor{} should not affect the probability.

\parbasic{General Approach}  We design our generative model $p(\ligsym | \recsym)$ by using a Markov decomposition:
\begin{equation}
    p(\ligsym | \recsym) = p(N, V, E, A | \recsym) = p(N | \recsym) \, \cdot \, p(V | N, \recsym) \, \cdot \, p(E| N, V, \recsym) \, \cdot \, p(A | N, V, E, \recsym)
    \label{eq:markov}
\end{equation}
In this paper, we focus on the first two terms on the right-hand side of the equation: the Number Distribution $p(N | \recsym)$ and the Vertex Distribution $p(V | N, \recsym)$.  The latter two terms may also be specified (e.g. see Appendix \ref{sec:edges_and_properties}), but are not the focus of our investigation in this paper.

\subsection{Two Flavours of EGNNs}
\label{sec:intermezzo}

In order to incorporate the relevant invariance properties, it will be helpful to use Equivariant Graph Neural Networks, also known as EGNNs \citep{satorras2021egnn}.  We now introduce two separate flavours of EGNNs, one which applies to the \receptor{} alone, and a second which applies to the combination of the \receptor{} and the \ligand{}.

\parbasic{\titlecap{\receptor{}} EGNN} \label{sec:receptor_egnn}
This is the standard EGNN which is described in \citep{satorras2021egnn}, applied to the \receptor{}.  As we are referring to the \receptor{}, we use hatted variables:
\begin{align}
    & \bh{m}^\ell_{ij} = \hat{\phi}_e(\bh{h}^\ell_i, \bh{h}^\ell_j, \|\bh{x}^\ell_i-\bh{x}^\ell_j\|^2, \|\bh{x}^0_i-\bh{x}^0_j\|^2, \bh{e}_{ij}, \{\bh{a}_k\})
    \hspace{0.65cm}
    \hat{b}_{ij}^\ell = \sigma(\hat{\phi}_b(\bh{m}_{ij}^\ell))
    \hspace{0.55cm}
    \bh{m}_i^\ell  = \sum_{j \in \hat{\eta}_i} \hat{b}_{ij}^\ell \bh{m}_{ij}^\ell
    \notag \\
    & \bh{x}^{\ell+1}_i=\bh{x}^\ell_i + \hat{\phi}_x(\bh{m}^\ell_{ij})
    \sum_{j\neq i} {\frac{(\bh{x}^\ell_i-\bh{x}^\ell_j)}{\|\bh{x}^\ell_i-\bh{x}^\ell_j\|+1}} 
    \hspace{1.7cm}
    \bh{h}^{\ell+1}_i=\bh{h}^{\ell}_i + \hat{\phi}_h(\bh{h}^\ell_i,\bh{m}^\ell_i)
    \label{eq:base_egnn}
\end{align}
The particular \titlecap{\receptor{}} EGNN is thus specified by the functions $\hat{\phi}_e, \hat{\phi}_b, \hat{\phi}_x, \hat{\phi}_h$.

\parbasic{\conditionalEGNN{}}
It is possible to design a joint EGNN on the \receptor{} and the \ligand{}, by constructing a single graph to capture both.  The main problem with this approach is that in many applications, the \receptor{} can be much larger than the \ligand{}; for example, in the molecular scenario the receptor is 1-2 orders of magnitude larger than the ligand.  As a result, this naive approach will lead to a situation in which the \ligand{} is ``drowned out'' by the \receptor{}, making it difficult to learn about the \ligand{}. We therefore take a different approach: we compute summary signatures of the \receptor{} based on the \titlecap{\receptor{}} EGNN, and use these as input to a \titlecap{\ligand{}} EGNN.  Our signatures will be based on the feature variables $\bh{h}^\ell_j$ from each layer $\ell = 1, \dots, \hat{L}$.  These variables are invariant to rigid body transformations by construction; furthermore, we can introduce permutation invariance by averaging, that is by using $\bh{h}^\ell_{av} = \Sigma_{j=1}^{\hat{N}} \bh{h}^\ell_j / \hat{N}$.
Thus, the \receptor{} at layer $\ell$ of the \ligand{}'s EGNN is summarized by the signature $\bh{g}^\ell$ which depends on both $\{ \bh{h}^\ell_{av} \}$, as well as the time $t$ of the normalizing flow ODE (to be introduced shortly in Section \ref{sec:vertex}):
\begin{equation}
    \bh{g}^0 = \phi^0_g \left( \bh{h}^1_{av}, \dots, \bh{h}^{\hat{L}}_{av}, t \right) \qquad \text{and} \qquad
    \bh{g}^\ell = \phi^\ell_g \left( \bh{g}^{\ell-1} \right) \quad \ell = 1, \dots, L
\end{equation}
The \conditionalEGNN{} is then specified by a form similar to the equations in (\ref{eq:base_egnn}), but in which the functions $\phi_e, \phi_b, \phi_x, \phi_h$ (now non-hatted, as we are referring to the \ligand{}) each have an additional dependence on the \receptor{} signatures $\{ \bh{g}^\ell \}_{\ell=1}^L$.  Further details are in Appendix \ref{sec:conditional_egnn}.

\subsection{The Number Distribution: $p(N | \recsym)$}
\label{sec:number}

\parbasic{Construction} Given the invariance conditions described in Equation (\ref{eq:invariance}), we propose the following distribution.  Let $\boldsymbol{\zeta}_N$ indicate a one-hot vector, where the index corresponding to $N$ is filled in with a $1$.  Based on the output of the \receptor{} EGNN, compute
\begin{equation}
    p(N | \recsym) = \boldsymbol{\zeta}_N^T F_2 \left( \Sigma_{i=1}^{\hat{N}} F_1 ( \bh{h}^L_i ) /  \hat{N} \right) 
\end{equation}
where $F_1$ and $F_2$ are multilayer perceptrons, and the last layer of $F_2$ is a softmax of size equal to the maximum number of atoms allowed. Since we use the $\bh{h}^L_i$ vectors (and not the $\bh{x}^L_i$ vectors), we have rigid motion invariance, as $T\bh{h}^L_i = \bh{h}^L_i$.  Since we use an average, we have permutation invariance.

\parbasic{Loss Function}  The loss function is straightforward: it is simply the negative log-likelihood of the number distribution, i.e. $L(\theta) = \mathbb{E}_{\ligsym, \recsym} [ -\log p(N | \recsym; \theta) ]$.

\subsection{The Vertex Distribution: $p(V | N, \recsym)$ via Continuous Normalizing Flows}
\label{sec:vertex}

\parbasic{General Notation} Define a vectorization operation on the vertex list $V$, which produces a vector $\mathbf{v}$; we refer to this as a \textit{vertex vector}.  Recall that $V = (\mathbf{v}_i)_{i=1}^N$ where $\mathbf{v}_i = (\mathbf{x}_i, \mathbf{h}_i)$.  Let
\begin{equation}
    \mathbf{x} = \texttt{concat}(\mathbf{x}_1, \dots, \mathbf{x}_N) 
    \hspace{0.8cm}
    \mathbf{h} = \texttt{concat}(\mathbf{h}_1, \dots, \mathbf{h}_N)
    \hspace{0.8cm}
    \mathbf{v} = \texttt{concat}(\mathbf{x}, \mathbf{h})
\end{equation}
The vertex vector $\mathbf{v} \in \mathbb{R}^{d_v^N}$ where $d_v^N = (d_h+3)N$. We denote the mapping from the vertex list $V$ to the vertex vector $\mathbf{v}$ as the vectorization operation $\texttt{vec}(\cdot)$: we write $\mathbf{v} = \texttt{vec}(V)$ and $V = \texttt{vec}^{-1}(\mathbf{v})$.
We have already described the action of rigid body transformations $T$ and permutations $\pi$ on the vertex list $V$ in Section \ref{sec:notation}.  It is easy to extend this to vertex vectors $\mathbf{v}$ using the \texttt{vec} operation; we have $T\mathbf{v} = \texttt{vec}( T\texttt{vec}^{-1}(\mathbf{v}) )$ and $\pi\mathbf{v} = \texttt{vec}( \pi\texttt{vec}^{-1}(\mathbf{v}) )$.

Given the above, it is sufficient for us to describe the distribution $p_{vec}(\mathbf{v} | \recsym)$
from which the vertex distribution $p(V | N, \recsym)$ follows directly, $p(V | N, \recsym) = p_{vec}(\texttt{vec}(V) | \recsym)$.  Note that we have suppressed $N$ in the condition in $p_{vec}(\cdot)$, as $\mathbf{v}$ is a vector of dimension $d_v^N$, so the $N$ dependence is already implicitly encoded.  

\parbasic{\ccmapping{} and Semi-Equivariance} Let $\gamma$ be a function which takes as input the \complex{} consisting of both the \ligand{} $\ligsym$ and \receptor{} $\recsym$, and outputs a new vertex list $V'$ for the \ligand{} $\ligsym$:
\begin{equation}
    V' = \gamma(\ligsym, \recsym)
\end{equation}
We refer to $\gamma$ as a \textit{\ccmapping{}}.  A rigid body transformation $T \in E(3)$ consists of a rotation and translation; let the rotation be denoted as $T_{rot}$.  Then we say that $\gamma$ is \textit{rotation semi-equivariant} if
\begin{equation}
    \gamma(T_{rot}\ligsym, T\recsym) = T_{rot}\gamma(\ligsym, \recsym) \quad \text{for all } T \in E(3)
\end{equation}
$\gamma$ is said to be \textit{permutation semi-equivariant} if
\begin{equation}
    \gamma(\pi \ligsym, \hat{\pi}\recsym) = \pi \gamma(\ligsym, \recsym) \quad \text{for all } \pi \in \mathbb{S}_N \text{ and } \hat{\pi} \in \mathbb{S}_{\hat{N}}
    \label{eq:gamma_permutation}
\end{equation}
Note in the definitions of both types of semi-equivariance, the differing roles played by the \ligand{} and \receptor{}; as the equivariant behaviour only applies to the \ligand{}, we have used the term semi-equivariance.

\parbasic{\conditionalFlow{}} Let $\gamma$ be a \ccmapping{}.  If $\mathbf{v}$ is a vertex vector, define $\ligsym_{\mathbf{v}}$ to be the graph $\ligsym$ with the vertex set replaced by $\texttt{vec}^{-1}(\mathbf{v})$. Then the following ordinary differential equation is referred to as a \textit{\conditionalFlow{}}:
\begin{equation}
    \frac{d\mathbf{u}}{dt} = \texttt{vec}\left( \gamma(\ligsym_{\mathbf{u}} , \recsym) \right), \quad \text{with } \mathbf{u}(0) = \mathbf{z}
\end{equation}
where the initial condition $\mathbf{z} \sim \mathcal{N}(0, \mathbf{I})$ is a Gaussian random vector of dimension $d_v^N$, and the ODE is run until $t=1$.  $\mathbf{u}(1)$ is thus the output of the \conditionalFlow{}.

\parbasic{Vertex Distributions with Appropriate Invariance} We now have the necessary ingredients to construct a distribution $p_{vec}(\mathbf{v} | \recsym)$ which yields a vertex distribution $p(V | N, \recsym)$ that satisfies the required invariance conditions.  The following is our main result (proof in Appendix \ref{sec:proof}):

\begin{theorem}[Invariant Vertex Distribution]
Let $\mathbf{u}(1)$ be the output of a \conditionalFlow{} specified by the \ccmapping{} $\gamma$.  Let the mean position of the \receptor{} be given by $\bh{x}_{av} = \frac{1}{\hat{N}} \sum_{i=1}^{\hat{N}} \bh{x}_i$, and define the following quantities
\begin{equation}
    \alpha = \frac{N}{N + \hat{N}}
    \quad
    \Omega_{\recsym} = 
    \begin{bmatrix}
        \mathbf{I}_{3N} - \frac{\alpha}{N} \mathbf{1}_{N \times N} \otimes \mathbf{I}_3 & \mathbf{0} \\
        \mathbf{0} & \mathbf{I}_{d_h N}
    \end{bmatrix}
    \quad
    \omega_{\recsym} = 
    \begin{bmatrix}
        -(1-\alpha) \mathbf{1}_{N \times 1} \otimes \bh{x}_{av} \\
        \mathbf{0}
    \end{bmatrix}
\end{equation}
where $\otimes$ indicates the Kronecker product.  Finally, let $\mathbf{v} = \Omega_{\recsym}^{-1} \left( \mathbf{u}(1) - \omega_{\recsym} \right)$.  Suppose that $\gamma$ is both rotation semi-equivariant and permutation semi-equivariant.  Then the resulting distribution on $\mathbf{v}$, that is $p_{vec}(\mathbf{v} | \recsym)$, yields a vertex distribution $p(V | N, \recsym) = p_{vec}(\texttt{vec}(V) | \recsym)$ that satisfies the invariance conditions in Equation (\ref{eq:invariance}).
\end{theorem}


\parbasic{Designing the \ccmapping{}} For the \ccmapping{}, we choose to use the \conditionalEGNN{}; that is, $V' = \gamma(\ligsym, \recsym)$ is given by $v'_i = (\mathbf{x}_i^L, \mathbf{h}_i^L)$, the output of the EGNN's final layer.
In practice, we use a slightly modified version of the foregoing, that is $v'_i = (\mathbf{x}_i^L-\mathbf{x}_i, \mathbf{h}_i^L)$, which we have found to converge well empirically. The rotation semi-equivariance of $\gamma$ (both versions - ordinary and modified) follows straightforwardly from the rotation semi-equivariance of EGNNs and the rotation invariance of the \receptor{} signatures $\{ \bh{g}^\ell \}_{\ell=1}^L$.  Similarly, the permutation semi-equivariance follows from the permutation equivariance of EGNNs and the permutation invariance of the \receptor{} signatures.

\parbasic{Loss Function}  The loss function and its optimization are implemented using standard techniques for continuous normalizing flows \citep{chen2018neural,grathwohl2018ffjord,chen2018continuous}.  If the feature vectors contain discrete variables, then techniques based on variational dequantization \citep{ho2019flow++} and argmax flows \citep{hoogeboom2021argmax} can be used, for ordinal and categorical features respectively.  This is parallel to the treatment in \citep{satorras2021flows}.

\section{Applications to Target-Aware Molecule Generation}
\label{sec:experiments}
\begin{table}[tb]
    \begin{subtable}{.5\linewidth}
      \centering
        \begin{tabular}{ |c||c|c|c|} 
            \hline
            \multicolumn{4}{|c|}{\textbf{Validity}} \\
            \hline
            \multicolumn{2}{|c|}{Ours} & \multicolumn{2}{|c|}{GraphBP} \\
            \hline
            \multicolumn{2}{|c|}{99.87\%} & \multicolumn{2}{|c|}{99.75\%} \\
            \hline
        \end{tabular}
    \end{subtable}%
    \begin{subtable}{.5\linewidth}
      \centering
        \begin{tabular}{ |c||c|c|c|} 
            \hline
            \multicolumn{4}{|c|}{$\mathbf{\Delta}$\textbf{Binding}} \\
            \hline
            \multicolumn{2}{|c|}{Ours} & \multicolumn{2}{|c|}{GraphBP} \\
            \hline
            \multicolumn{2}{|c|}{35.7\%} & \multicolumn{2}{|c|}{22.76\%} \\
            \hline
        \end{tabular}
    \end{subtable}
    \caption{Comparison of molecule validity and $\Delta$Binding between proposed method and GraphBP.}
    \label{tab:validity_binding}
    \vspace{-0mm}
\end{table}

We now apply the theory we have developed to the problem of target-aware structure based molecule generation, in which the \receptor{} plays the role of the receptor, while the \ligand{} plays the role of the ligand; vertices are atoms and edges are bonds.  We train our model on the CrossDocked2020 dataset \citep{francoeur2020three}.  Further details are included in Appendix \ref{sec:experimental_setup}.

The validity is defined as the percentage of molecules that are chemically valid, in the sense that they can be sanitized by RDKit, see \citep{Landrum2016RDKit2016_09_4}. As shown in Table \ref{tab:validity_binding}, our model produces ligands with a validity of 99.87\%, surpassing the baseline, GraphBP \citep{liu2022generating}.  $\Delta$Binding measures the percentage of generated molecules that have higher predicted binding affinity to the target binding site than the corresponding reference molecule.  As can be seen in Table \ref{tab:validity_binding}, our method attains $\Delta$Binding = 35.7\% vs. 22.76\% for GraphBP, which corresponds to a relative improvement of 56.81\%.  Further details are included in Appendix \ref{sec:detailed_experimental_results}.

\newpage
\bibliography{iclr2023_workshop}
\bibliographystyle{iclr2023_workshop}

\appendix
\section{Related Work}
\label{sec:related_work}
Continuous normalizing flows originated in the work of \citet{chen2018continuous} and have been extended in many ways, for example \citep{grathwohl2018ffjord,zhang2022pnode,onken2021ot,ghosh2020steer}. \citet{kohler2019equivariant} and \citet{equi_hamiltonianflows} presented equivariant normalizing flows which respect natural symmetries.  \citet{NEURIPS2019_1e44fdf9} introduced graph normalizing flows to obtain generative models of graph structures. \citet{kohler2020equivariant} and \citet{satorras2021flows} constructed equivariant graph normalizing flows, by incorporating equivariant graph neural networks into an ODE framework to obtain invertible equivariant functions.  A variety of works have tackled the problem of receptor-aware ligand generation, including methods based on VAEs, e.g. \citep{ragoza2022generating}; autoregressive models, e.g. \citep{luo20213d, liu2022generating,peng2022pocket2mol,drotar2021structure}; diffusion, e.g. \citep{schneuing2022structure, lin2022diffbp}; and reinforcement learning, e.g. \citep{li2021structure,thomas2021comparison,fialkova2021libinvent,fu2022reinforced}.

\section{The \conditionalEGNN}
\label{sec:conditional_egnn}
The \receptor{} at layer $\ell$ of the \ligand{}'s EGNN is summarized by the signature $\bh{g}^\ell$ which depends on both $\{ \bh{h}^\ell_{av} \}$ and the ODE time $t$:
\begin{equation}
    \bh{g}^0 = \phi^0_g \left( \bh{h}^1_{av}, \dots, \bh{h}^{\hat{L}}_{av}, t \right) \qquad \text{and} \qquad
    \bh{g}^\ell = \phi^\ell_g \left( \bh{g}^{\ell-1} \right) \quad \ell = 1, \dots, L
\end{equation}
Note that there are $L+1$ separate functions $\{ \phi^\ell_g \}_{\ell=0}^L$.

These invariant \receptor{} signatures $\{ \bh{g}^\ell \}_{\ell=1}^L$ are then naturally incorporated into the \conditionalEGNN{} as follows:
\begin{align}
    & \mathbf{m}^\ell_{ij} = \phi_e\left(\mathbf{h}^\ell_i,\mathbf{h}^\ell_j,  \|\mathbf{x}^\ell_i-\mathbf{x}^\ell_j\|^2, \|\mathbf{x}^0_i-\mathbf{x}^0_j\|^2 , \bh{g}^\ell, t \right)
    \hspace{0.5cm}
    \quad b_{ij}^\ell = \sigma(\phi_b(\mathbf{m}_{ij}^\ell, \bh{g}^\ell ))
    \hspace{0.5cm}
    \mathbf{m}^\ell_i  = \sum_{j=1}^N b^\ell_{ij} \mathbf{m}^\ell_{ij}
    \notag \\
    & \mathbf{x}^{\ell+1}_i= \mathbf{x}^\ell_i + \left( \sum_{j\neq i} {\frac{(\mathbf{x}^\ell_i-\mathbf{x}^\ell_j)}{\|\mathbf{x}^\ell_i-\mathbf{x}^\ell_j\|+1}} \right) \phi_x(\mathbf{m}^\ell_{ij} , \bh{g}^\ell )
    \hspace{0.7cm}
    \mathbf{h}^{\ell+1}_i = \mathbf{h}^{\ell}_i + \phi_h(\mathbf{h}^\ell_i, \mathbf{m}^\ell_i , \bh{g}^\ell )
\end{align}
The particular \conditionalEGNN{} is thus specified by the functions $\{ \phi^\ell_g \}_{\ell=0}^L, \phi_e, \phi_b, \phi_x, \phi_h$.

\section{Proof of the Invariant Vertex Distribution Theorem}
\label{sec:proof}
In this section, we prove the Invariant Vertex Distribution Theorem presented in Section \ref{sec:vertex}.  We begin with two lemmata, after which the proof of the theorem is presented.

\begin{lemma}
\label{lemma:first_transformation}
Let the mean position of the \receptor{} be given by $\bh{x}_{av} = \frac{1}{\hat{N}} \sum_{i=1}^{\hat{N}} \bh{x}_i$, and define the following quantities
\begin{equation*}
    \alpha = \frac{N}{N + \hat{N}} \quad \quad
    \Omega_{\recsym} = 
    \begin{bmatrix}
        \mathbf{I}_{3N} - \frac{\alpha}{N} \mathbf{1}_{N \times N} \otimes \mathbf{I}_3 & \mathbf{0} \\
        \mathbf{0} & \mathbf{I}_{d_h N}
    \end{bmatrix}
    \quad \quad
    \omega_{\recsym} = 
    \begin{bmatrix}
        -(1-\alpha) \mathbf{1}_{N \times 1} \otimes \bh{x}_{av} \\
        \mathbf{0}
    \end{bmatrix}
\end{equation*}
where $\otimes$ indicates the Kronecker product.  Given the following mapping:
\begin{equation}
    \mathbf{v} = \Omega_{\recsym}^{-1} \left( \mathbf{u} - \omega_{\recsym} \right)
\end{equation}
Let the inverse mapping be denoted by $\Gamma_{\recsym}^1$, i.e. $\mathbf{u} = \Gamma_{\recsym}^1(\mathbf{v})$.  For any rigid transformation $T \in E(3)$, which consists of both a rotation and a translation, denote the transformation consisting only of the rotation of $T$ as $T_{rot} \in O(3)$.  Then
\begin{equation*}
    \Gamma_{T\recsym}^1(T\mathbf{v}) = T_{rot} \Gamma_{\recsym}^1(\mathbf{v}).
\end{equation*}
Furthermore, for any permutations $\pi \in \mathbb{S}_N$ and $\hat{\pi} \in \mathbb{S}_{\hat{N}}$, then 
\begin{equation*}
    \Gamma_{\hat{\pi}\recsym}^1(\pi\mathbf{v}) = \pi \Gamma_{\recsym}^1(\mathbf{v}).
\end{equation*}
\end{lemma}

\textbf{\textit{Proof:}} The mapping $\mathbf{u} = \Gamma_{\recsym}^1(\mathbf{v})$ is given by
\begin{equation}
    \mathbf{u} = \Omega_{\recsym} \mathbf{v} + \omega_{\recsym}
\end{equation}
Let us denote the parts of $\mathbf{u}$ corresponding to the coordinates and the features as $\mathbf{x}^\mathbf{u}$ and $\mathbf{h}^\mathbf{u}$, respectively; and use similar notation for $\mathbf{v}$.  Then we have that
\begin{equation}
    \mathbf{h}^\mathbf{u} = \mathbf{h}^\mathbf{v}
    \label{eq:h_trans}
\end{equation}
and
\begin{equation}
    \mathbf{x}^\mathbf{u} = \left( \mathbf{I}_{3N} - \frac{\alpha}{N} \mathbf{1}_{N \times N} \otimes \mathbf{I}_3 \right) \mathbf{x}^\mathbf{v} -(1-\alpha) \mathbf{1}_{N \times 1} \otimes \bh{x}_{av}
\end{equation}
Breaking down this last equation by vertex gives
\begin{align}
    \mathbf{x}_i^\mathbf{u}
    & = \mathbf{x}_i^\mathbf{v} - \frac{\alpha}{N} \sum_{j=1}^N \mathbf{x}_j^\mathbf{v} -(1-\alpha) \bh{x}_{av} \notag \\
    & = \mathbf{x}_i^\mathbf{v} - \left( \alpha \mathbf{x}_{av}^\mathbf{v} + (1-\alpha) \bh{x}_{av} \right) \notag \\
    & = \mathbf{x}_i^\mathbf{v} - \bar{\mathbf{x}}^\mathbf{v}
    \label{eq:per_coord_trans}
\end{align}
where $\mathbf{x}_{av}^\mathbf{v}$ is the average coordinate position of $\mathbf{x}^\mathbf{v}$, and $\bar{\mathbf{x}}^\mathbf{v}$ indicates the average of all vertices in the entire \complex{}, i.e. taking both the \ligand{} and the \receptor{} together.

Now, let us examine what happens when we apply the rigid transformation $T$ to both $\mathbf{v}$ and the \receptor{} $\recsym$; that is, let us examine
\begin{equation}
    \tilde{\mathbf{u}} = \Gamma_{T\recsym}^1(T\mathbf{v})
\end{equation}
In the case of the features $\mathbf{h}$, they are invariant by design; thus 
\begin{align}
    \mathbf{h}^{\tilde{\mathbf{u}}} 
    & = \mathbf{h}^\mathbf{Tv} \notag \\
    & = \mathbf{h}^\mathbf{v} \notag \\
    & = \mathbf{h}^\mathbf{u}
    \label{eq:u_trans_h}
\end{align}
where the last line follows from Equation (\ref{eq:h_trans}).
In the case of the coordinates, the transformation is as follows:
\begin{equation}
    \mathbf{x}_i^{T\mathbf{v}} = R\mathbf{x}_i^\mathbf{v} + t
\end{equation}
where $R \in O(3)$ is the rotation matrix, and $t \in \mathbb{R}^3$ the translation vector, corresponding to rigid motion $T$.  As we apply $T$ to the \receptor{} $\recsym$, this has the effect of applying this transformation to each of the \receptor{} vertices, and hence to their mean and the mean of the entire \complex{}:
\begin{equation}
    \bh{x}_{av}^{T\recsym} = R \bh{x}_{av}^{\recsym} + t \quad \Rightarrow \quad \bar{\mathbf{x}}^\mathbf{Tv} = R \bar{\mathbf{x}}^\mathbf{v} + t
\end{equation}
Thus, following Equation (\ref{eq:per_coord_trans}), and substituting $T\mathbf{v}$ and $T\recsym$ in place of $\mathbf{v}$ and $\recsym$, we get
\begin{align}
    \mathbf{x}_i^{\tilde{\mathbf{u}}}
    & = \mathbf{x}_i^{T\mathbf{v}} - \bar{\mathbf{x}}^{T\mathbf{v}} \notag \\
    & = R\mathbf{x}_i^\mathbf{v} + t - (R\bar{\mathbf{x}}^\mathbf{v} + t ) \notag \\
    & = R \left( \mathbf{x}_i^\mathbf{v} - \bar{\mathbf{x}}^\mathbf{v} \right) \notag \\
    & = R \mathbf{x}_i^\mathbf{u}
    \label{eq:u_trans_x}
\end{align}
Combining Equations (\ref{eq:u_trans_h}) and (\ref{eq:u_trans_x}), we have that
\begin{equation}
    \tilde{\mathbf{u}} = T_{rot} \mathbf{u}
\end{equation}
Since $\mathbf{u} = \Gamma_{\recsym}^1(\mathbf{v})$ and $\tilde{\mathbf{u}} = \Gamma_{T\recsym}^1(T\mathbf{v})$, we have shown that $\Gamma_{T\recsym}^1(T\mathbf{v}) = T_{rot} \Gamma_{\recsym}^1(\mathbf{v})$, as desired. 

In the case of the permutations, let us now set
\begin{equation}
    \tilde{\mathbf{u}} = \Gamma_{\hat{\pi}\recsym}^1(\pi\mathbf{v})
\end{equation}
It is easy to see that $\hat{\pi}$ has no effect; the only place the \receptor{} enters is through the quantities $\hat{N}$ and $\bh{x}_{av}$, both of which are permutation-invariant.  For the features, we now have
\begin{align}
    \mathbf{h}^{\tilde{\mathbf{u}}} 
    & = \mathbf{h}^\mathbf{\pi v} \notag \\
    & = \pi \mathbf{h}^\mathbf{v} \notag \\
    & = \pi \mathbf{h}^\mathbf{u}
\end{align}
That is, the features are simply reordered according to $\pi$.  With regard to the coordinates, we have that
\begin{align}
    \mathbf{x}_i^{\tilde{\mathbf{u}}}
    & = \mathbf{x}_i^{\pi\mathbf{v}} - \bar{\mathbf{x}}^{\pi\mathbf{v}} \notag \\
    & = \mathbf{x}_{\pi(i)}^\mathbf{v} - \bar{\mathbf{x}}^\mathbf{v} \notag \\
    & = \mathbf{x}_{\pi(i)}^\mathbf{u}
\end{align}
The coordinates are also therefore simply reordered according to $\pi$.  Summarizing, we have that
\begin{equation}
    \tilde{\mathbf{u}} = \pi \mathbf{u}
\end{equation}
This is exactly equal to 
\begin{equation*}
    \Gamma_{\hat{\pi}\recsym}^1(\pi\mathbf{v}) = \pi \Gamma_{\recsym}^1(\mathbf{v})
\end{equation*}
which concludes the proof. \qedsymbol

\begin{lemma}
\label{lemma:second_transformation}
Let $\mathbf{u}(1)$ be the output of a \conditionalFlow{} specified by the \ccmapping{} $\gamma$ which is rotation semi-equivariant and permutation semi-equivariant.  This \conditionalFlow{} maps the initial condition $\mathbf{z}$ to $\mathbf{u}(1)$; let the inverse mapping be denoted by $\Gamma_{\recsym}^2$, i.e. $\mathbf{z} = \Gamma_{\recsym}^2(\mathbf{u}(1))$.  Then
\begin{equation*}
    \Gamma_{T\recsym}^2(T_{rot} \mathbf{u}) = T_{rot} \Gamma_{\recsym}^2(\mathbf{u})
\end{equation*}
Furthermore, for any permutations $\pi \in \mathbb{S}_N$ and $\hat{\pi} \in \mathbb{S}_{\hat{N}}$, then
\begin{equation*}
    \Gamma_{\hat{\pi}\recsym}^2(\pi \mathbf{u}) = \pi \Gamma_{\recsym}^2(\mathbf{u})
\end{equation*}
\end{lemma}

\textbf{\textit{Proof:}} Our first goal is to show that $\Gamma_{T\recsym}^2(T_{rot} \mathbf{u}) = T_{rot} \Gamma_{\recsym}^2(\mathbf{u})$.  Let us define $F_{\recsym}$ to be the inverse of $\Gamma_{\recsym}^2$, and let $\mathbf{u} = F_{\recsym}(\mathbf{z})$.  Then
\begin{align}
    \Gamma_{T\recsym}^2(T_{rot} \mathbf{u}) = T_{rot} \Gamma_{\recsym}^2(\mathbf{u})
    & \quad \Leftrightarrow \quad F_{T\recsym}^{-1}(T_{rot} F_{\recsym}(\mathbf{z})) = T_{rot} F_{\recsym}^{-1}(F_{\recsym}(\mathbf{z})) \notag \\
    & \quad \Leftrightarrow \quad F_{T\recsym}^{-1}(T_{rot} F_{\recsym}(\mathbf{z})) = T_{rot} \mathbf{z} \notag \\
    & \quad \Leftrightarrow \quad F_{T\recsym}( T_{rot} \mathbf{z} ) = T_{rot} F_{\recsym}(\mathbf{z})
\end{align}
Thus, it is sufficient to show that $F_{T\recsym}( T_{rot} \mathbf{z} ) = T_{rot} F_{\recsym}(\mathbf{z})$.  For convenience, we shall set
\begin{equation}
    \mathbf{u}(1) = F_{\recsym}(\mathbf{z}) \quad \text{and} \quad \tilde{\mathbf{u}}(1) = F_{T\recsym}( T_{rot} \mathbf{z} )
    \label{eq:u_cond1}
\end{equation}
In this case, $\mathbf{u}(1)$ is defined by the ODE
\begin{equation}
    \frac{d\mathbf{u}}{dt} = \texttt{vec}\left( \gamma(\ligsym_{\mathbf{u}} , \recsym) \right) \quad \text{with } \mathbf{u}(0) = \mathbf{z}
    \label{eq:u_flow1}
\end{equation}
whereas $\tilde{\mathbf{u}}(1)$ is defined by the ODE
\begin{equation}
    \frac{d\tilde{\mathbf{u}}}{dt} = \texttt{vec}\left( \gamma(\ligsym_{\tilde{\mathbf{u}}} , T\recsym) \right) \quad \text{with } \tilde{\mathbf{u}}(0) = T_{rot} \mathbf{z}
    \label{eq:utilde_flow1}
\end{equation}
Now, let us define $\Breve{\mathbf{u}}(t) = T_{rot}^{-1} \tilde{\mathbf{u}}(t)$, so that $\tilde{\mathbf{u}}(t) = T_{rot} \Breve{\mathbf{u}}(t)$.  In this case, we have that:
\begin{enumerate}
    \item $\Breve{\mathbf{u}}(0) = T_{rot}^{-1} \tilde{\mathbf{u}}(0) = T_{rot}^{-1} T_{rot} \mathbf{z} = \mathbf{z}$.
    \item $\frac{d\tilde{\mathbf{u}}}{dt} = T_{rot} \frac{d\Breve{\mathbf{u}}}{dt}$.
    \item $\texttt{vec}\left( \gamma(\ligsym_{\tilde{\mathbf{u}}} , T\recsym) \right) = \texttt{vec}\left( \gamma(\ligsym_{T_{rot}\Breve{\mathbf{u}}} , T\recsym) \right) = T_{rot} \texttt{vec}\left( \gamma(\ligsym_{\Breve{\mathbf{u}}} , \recsym) \right)$, where the last equality is from the definition of rotation semi-equivariance of $\gamma$.
\end{enumerate}
Plugging the above three results into the flow for $\tilde{\mathbf{u}}$ in Equation (\ref{eq:utilde_flow1}) yields
\begin{align}
    & T_{rot} \frac{d\Breve{\mathbf{u}}}{dt}  = T_{rot} \texttt{vec}\left( \gamma(\ligsym_{\Breve{\mathbf{u}}} , \recsym) \right) \quad \text{with} \quad \Breve{\mathbf{u}}(0) = \mathbf{z} \notag \\
    & \Rightarrow \quad \frac{d\Breve{\mathbf{u}}}{dt}  = \texttt{vec}\left( \gamma(\ligsym_{\Breve{\mathbf{u}}} , \recsym) \right) \quad \text{with} \quad \Breve{\mathbf{u}}(0) = \mathbf{z}
\end{align}
But this is precisely identical to the flow described in Equation (\ref{eq:u_flow1}); thus, we have that
\begin{equation}
    \Breve{\mathbf{u}}(t) = \mathbf{u}(t) \quad \text{for all } t
\end{equation}
But $\tilde{\mathbf{u}}(t) = T_{rot} \Breve{\mathbf{u}}(t)$ so that $\tilde{\mathbf{u}}(t) = T_{rot} \mathbf{u}(t)$, and in particular $\tilde{\mathbf{u}}(1) = T_{rot} \mathbf{u}(1)$.  Comparing with Equation (\ref{eq:u_cond1}) completes rigid motion part of the proof.

Let us now turn to permutations; the proof is similar, but we repeat it in full for completeness.  Our goal is to show that $\Gamma_{\hat{\pi}\recsym}^2(\pi \mathbf{u}) = \pi \Gamma_{\recsym}^2(\mathbf{u})$.  Let us define $F_{\recsym}$ to be the inverse of $\Gamma_{\recsym}^2$, and let $\mathbf{u} = F_{\recsym}(\mathbf{z})$.  Then
\begin{align}
    \Gamma_{\hat{\pi}\recsym}^2(\pi \mathbf{u}) = \pi \Gamma_{\recsym}^2(\mathbf{u})
    & \quad \Leftrightarrow \quad F_{\hat{\pi}\recsym}^{-1}(\pi F_{\recsym}(\mathbf{z})) = \pi F_{\recsym}^{-1}(F_{\recsym}(\mathbf{z})) \notag \\
    & \quad \Leftrightarrow \quad F_{\hat{\pi}\recsym}^{-1}(\pi F_{\recsym}(\mathbf{z})) = \pi \mathbf{z} \notag \\
    & \quad \Leftrightarrow \quad F_{\hat{\pi}\recsym}( \pi \mathbf{z} ) = \pi F_{\recsym}(\mathbf{z})
\end{align}
Thus, it is sufficient to shows that $F_{\hat{\pi}\recsym}( \pi \mathbf{z} ) = \pi F_{\recsym}(\mathbf{z})$.  For convenience, we shall set
\begin{equation}
    \mathbf{u}(1) = F_{\recsym}(\mathbf{z}) \quad \text{and} \quad \tilde{\mathbf{u}}(1) = F_{\hat{\pi}\recsym}( \pi \mathbf{z} )
    \label{eq:u_cond}
\end{equation}
In this case, $\mathbf{u}(1)$ is defined by the ODE
\begin{equation}
    \frac{d\mathbf{u}}{dt} = \texttt{vec}\left( \gamma(\ligsym_{\mathbf{u}} , \recsym) \right) \quad \text{with } \mathbf{u}(0) = \mathbf{z}
    \label{eq:u_flow}
\end{equation}
whereas $\tilde{\mathbf{u}}(1)$ is defined by the ODE
\begin{equation}
    \frac{d\tilde{\mathbf{u}}}{dt} = \texttt{vec}\left( \gamma(\ligsym_{\tilde{\mathbf{u}}} , \hat{\pi}\recsym) \right) \quad \text{with } \tilde{\mathbf{u}}(0) = \pi \mathbf{z}
    \label{eq:utilde_flow}
\end{equation}
Now, let us define $\Breve{\mathbf{u}}(t) = \pi^{-1} \tilde{\mathbf{u}}(t)$, so that $\tilde{\mathbf{u}}(t) = \pi \Breve{\mathbf{u}}(t)$.  In this case, we have that:
\begin{enumerate}
    \item $\Breve{\mathbf{u}}(0) = \pi^{-1} \tilde{\mathbf{u}}(0) = \pi^{-1} \pi \mathbf{z} = \mathbf{z}$.
    \item $\frac{d\tilde{\mathbf{u}}}{dt} = \pi \frac{d\Breve{\mathbf{u}}}{dt}$.
    \item $\texttt{vec}\left( \gamma(\ligsym_{\tilde{\mathbf{u}}} , \hat{\pi}\recsym) \right) = \texttt{vec}\left( \gamma(\ligsym_{\pi\Breve{\mathbf{u}}} , \hat{\pi}\recsym) \right) = \pi \texttt{vec}\left( \gamma(\ligsym_{\Breve{\mathbf{u}}} , \recsym) \right)$, where the last equality is from the definition of permutation semi-equivariance of $\gamma$.
\end{enumerate}
Plugging the above three results into the flow for $\tilde{\mathbf{u}}$ in Equation (\ref{eq:utilde_flow}) yields
\begin{align}
    & \pi \frac{d\Breve{\mathbf{u}}}{dt}  = \pi \texttt{vec}\left( \gamma(\ligsym_{\Breve{\mathbf{u}}} , \recsym) \right) \quad \text{with} \quad \Breve{\mathbf{u}}(0) = \mathbf{z} \notag \\
    & \Rightarrow \quad \frac{d\Breve{\mathbf{u}}}{dt}  = \texttt{vec}\left( \gamma(\ligsym_{\Breve{\mathbf{u}}} , \recsym) \right) \quad \text{with} \quad \Breve{\mathbf{u}}(0) = \mathbf{z}
\end{align}
But this is precisely identical to the flow described in Equation (\ref{eq:u_flow}); thus, we have that
\begin{equation}
    \Breve{\mathbf{u}}(t) = \mathbf{u}(t) \quad \text{for all } t
\end{equation}
But $\tilde{\mathbf{u}}(t) = \pi \Breve{\mathbf{u}}(t)$ so that $\tilde{\mathbf{u}}(t) = \pi \mathbf{u}(t)$, and in particular $\tilde{\mathbf{u}}(1) = \pi \mathbf{u}(1)$.  Comparing with Equation (\ref{eq:u_cond}) completes the proof. \qedsymbol

\begin{theorem}
Let $\mathbf{u}(1)$ be the output of a \conditionalFlow{} specified by the \ccmapping{} $\gamma$.  Let the mean position of the \receptor{} be given by $\bh{x}_{av} = \frac{1}{\hat{N}} \sum_{i=1}^{\hat{N}} \bh{x}_i$, and define the following quantities
\begin{equation}
    \alpha = \frac{N}{N + \hat{N}} \quad \quad
    \Omega_{\recsym} = 
    \begin{bmatrix}
        \mathbf{I}_{3N} - \frac{\alpha}{N} \mathbf{1}_{N \times N} \otimes \mathbf{I}_3 & \mathbf{0} \\
        \mathbf{0} & \mathbf{I}_{d_h N}
    \end{bmatrix}
    \quad \quad
    \omega_{\recsym} = 
    \begin{bmatrix}
        -(1-\alpha) \mathbf{1}_{N \times 1} \otimes \bh{x}_{av} \\
        \mathbf{0}
    \end{bmatrix}
\end{equation}
where $\otimes$ indicates the Kronecker product.  Finally, let 
\begin{equation}
    \mathbf{v} = \Omega_{\recsym}^{-1} \left( \mathbf{u}(1) - \omega_{\recsym} \right)
    \label{eq:final_transformation}
\end{equation}

Suppose that $\gamma$ is both rotation semi-equivariant and permutation semi-equivariant.  Then the resulting distribution on $\mathbf{v}$, that is $p_{vec}(\mathbf{v} | \recsym)$, yields a vertex distribution $p(V | N, \recsym) = p_{vec}(\texttt{vec}(V) | \recsym)$ that satisfies the invariance conditions in Equation (\ref{eq:invariance}).
\end{theorem}

\textbf{\textit{Proof:}} The \conditionalFlow{} maps from the Gaussian random variable $\mathbf{z}$ to the variable $\mathbf{u}(1)$.  As this flow is a normalizing flow, it is invertible, so let us denote the inverse mapping by $\Gamma^2$:
\begin{equation}
    \mathbf{z} = \Gamma_{\recsym}^2(\mathbf{u}(1))
\end{equation}
Note that the dependence on the \receptor{} $\recsym$ is denoted using a subscript, as the invertibility does not apply to the \receptor{}, but only to the \ligand{}.  Equation (\ref{eq:final_transformation}) maps from the variable $\mathbf{u}(1)$ to the variable $\mathbf{v}$; let us denote its inverse mapping by $\Gamma^1$:
\begin{equation}
    \mathbf{u}(1) = \Gamma_{\recsym}^1(\mathbf{v})
\end{equation}
In this case, we have that
\begin{equation}
    \mathbf{z} = \Gamma_{\recsym}^2(\Gamma_{\recsym}^1(\mathbf{v})) \equiv \Gamma_{\recsym}(\mathbf{v})
    \label{eq:v_to_z}
\end{equation}

Now, our goal is to show that the following condition holds:
\begin{equation}
    p(TV | N, T\recsym) = p(V | N, \recsym) \hspace{0.5cm} \text{for } T \in E(3)
\end{equation}
Using the $p_{vec}$ notation, this translates to
\begin{equation}
    p_{vec}(T\mathbf{v} | T\recsym) = p_{vec}(\mathbf{v} | \recsym)
    \label{eq:invariance_with_vec}
\end{equation}
Now, from Equation (\ref{eq:v_to_z}), the fact that $\Gamma$ is invertible, and the change of variables formula, we have that
\begin{equation}
    p_{vec}(\mathbf{v} | \recsym) = p_{\mathbf{z}}(\Gamma_{\recsym}(\mathbf{v})) |\det J_{\Gamma_{\recsym}}(\mathbf{v}) |
\end{equation}
where $p_{\mathbf{z}}(\cdot)$ is the Gaussian distribution from $\mathbf{z}$ is sampled; and $J_{\Gamma_{\recsym}}(\cdot)$ is the Jacobian of $\Gamma_{\recsym}(\cdot)$.  Since $\Gamma_{\recsym} = \Gamma_{\recsym}^2 \circ \Gamma_{\recsym}^1$, this can be expanded as
\begin{equation}
    p_{vec}(\mathbf{v} | \recsym) = p_{\mathbf{z}}( \Gamma_{\recsym}^2(\Gamma_{\recsym}^1(\mathbf{v})) ) |\det J_{\Gamma_{\recsym}^2}(\Gamma_{\recsym}^1(\mathbf{v})) | |\det J_{\Gamma_{\recsym}^1}(\mathbf{v}) |
\end{equation}
using the chain rule, and the fact that determinant of a product is the product of determinants.  Plugging this into Equation (\ref{eq:invariance_with_vec}), we must show that
\begin{align}
    & p_{\mathbf{z}}( \Gamma_{T\recsym}^2(\Gamma_{T\recsym}^1(T\mathbf{v})) ) |\det J_{\Gamma_{T\recsym}^2}(\Gamma_{T\recsym}^1(T\mathbf{v})) | |\det J_{\Gamma_{T\recsym}^1}(T\mathbf{v}) |  \notag \\
    & \hspace{1.0cm} = \quad p_{\mathbf{z}}( \Gamma_{\recsym}^2(\Gamma_{\recsym}^1(\mathbf{v})) ) |\det J_{\Gamma_{\recsym}^2}(\Gamma_{\recsym}^1(\mathbf{v})) | |\det J_{\Gamma_{\recsym}^1}(\mathbf{v}) |
    \hspace{0.5cm} \text{for } T \in E(3)
    \label{eq:to_prove}
\end{align}
A rigid transformation $T \in E(3)$ consists of both a rotation and a translation.  For brevity, denote the transformation consisting only of the rotation of $T$ as $T_{rot} \in O(3)$.  Now, from Lemma \ref{lemma:first_transformation}, we have that
\begin{equation}
    \Gamma_{T\recsym}^1(T\mathbf{v}) = T_{rot} \Gamma_{\recsym}^1(\mathbf{v})
    \label{eq:lemma_repr_a}
\end{equation}
From Lemma \ref{lemma:second_transformation}, we have that
\begin{equation}
    \Gamma_{T\recsym}^2(T_{rot} \mathbf{u}) = T_{rot} \Gamma_{\recsym}^2(\mathbf{u})
    \label{eq:lemma_repr_b}
\end{equation}
Combining Equations (\ref{eq:lemma_repr_a}) and (\ref{eq:lemma_repr_b}) gives that
\begin{align}
    p_{\mathbf{z}}( \Gamma_{T\recsym}^2(\Gamma_{T\recsym}^1(T \mathbf{v})) )
    & = p_{\mathbf{z}}( \Gamma_{T\recsym}^2(T_{rot}\Gamma_{\recsym}^1(\mathbf{v})) ) \notag \\
    & = p_{\mathbf{z}}( T_{rot} \Gamma_{\recsym}^2(\Gamma_{\recsym}^1(\mathbf{v})) ) \notag \\
    & = p_{\mathbf{z}}( \Gamma_{\recsym}^2(\Gamma_{\recsym}^1(\mathbf{v})) )
    \label{eq:distribution_result_1}
\end{align}
where the last line follows from the rotation invariance of the Gaussian distribution.

Note that
\begin{align}
    J_{\Gamma_{T\recsym}^1}(T\mathbf{v})
    & = \frac{\partial}{\partial \mathbf{v}} \left( \Gamma_{T\recsym}^1 (T\mathbf{v}) \right) \notag \\
    & = \frac{\partial}{\partial \mathbf{v}} \left( T_{rot}\Gamma_{\recsym}^1(\mathbf{v}) \right) \notag \\
    & = T_{rot} \frac{\partial}{\partial \mathbf{v}} \left( \Gamma_{\recsym}^1(\mathbf{v}) \right) \notag \\
    & = T_{rot} J_{\Gamma_{\recsym}^1}(\mathbf{v})
    \label{eq:JGamma1}
\end{align}
Now, $T_{rot}$ can be represented by the $d_v^N \times d_v^N$ block diagonal matrix given by
\begin{equation}
    T_{rot} =
    \begin{bmatrix}
        \mathbf{1}_{N \times 1} \otimes R & 0 \\
        0 & \mathbf{I}_{d_h N}
    \end{bmatrix}
\end{equation}
where $R \in O(3)$, the top-left block corresponds to the coordinates $\mathbf{x}$ and the bottom-right block corresponds to the feature $\mathbf{h}$.  Thus,
\begin{align}
    \det( J_{\Gamma_{T\recsym}^1}(T\mathbf{v}) )
    & = \det( T_{rot} J_{\Gamma_{\recsym}^1}(\mathbf{v}) ) \notag \\
    & = \det( T_{rot} ) \det( J_{\Gamma_{\recsym}^1}(\mathbf{v}) ) \notag \\
    & = \det(R)^N \det(\mathbf{I}_{d_h N}) \det( J_{\Gamma_{\recsym}^1}(\mathbf{v}) ) \notag \\
    & = \pm \det( J_{\Gamma_{\recsym}^1}(\mathbf{v}) )
    \label{eq:jacobian_result_1}
\end{align}
where the second line follows from the fact that the determinant of a product is the product of determinants; the third line from the fact that the determinant of a block diagonal matrix is the product of the determinants of the blocks; and the fourth line from the fact that the determinant of a rotation matrix is $\pm 1$.

To simplify $J_{\Gamma_{T\recsym}^2}(\Gamma_{T\recsym}^1(T\mathbf{v}))$, note that
\begin{align}
    \Gamma_{T\recsym}^2(\mathbf{u})
    & = \Gamma_{T\recsym}^2(T_{rot} T_{rot}^{-1} \mathbf{u}) \notag \\
    & = T_{rot} \Gamma_{\recsym}^2( T_{rot}^{-1} \mathbf{u})
\end{align}
where we have used Equation (\ref{eq:lemma_repr_b}).  Thus,
\begin{align}
    J_{\Gamma_{T\recsym}^2}(\mathbf{u})
    & = \frac{\partial}{\partial \mathbf{u}} \left( \Gamma_{T\recsym}^2(\mathbf{u}) \right) \notag \\
    & = \frac{\partial}{\partial \mathbf{u}} \left( T_{rot} \Gamma_{\recsym}^2( T_{rot}^{-1} \mathbf{u}) \right) \notag \\
    & = T_{rot} J_{\Gamma_{T\recsym}^2}( T_{rot}^{-1} \mathbf{u} ) T_{rot}^{-1}
    \label{eq:jac_simp}
\end{align}
We wish to plug in $\mathbf{u} = \Gamma_{T\recsym}^1(T\mathbf{v})$.  Note that from Equation (\ref{eq:lemma_repr_a}), $\Gamma_{T\recsym}^1(T\mathbf{v}) = T_{rot} \Gamma_{\recsym}^1(\mathbf{v})$.  Thus,
\begin{align}
    J_{\Gamma_{T\recsym}^2}(\Gamma_{T\recsym}^1(T\mathbf{v}))
    & = J_{\Gamma_{T\recsym}^2}( T_{rot} \Gamma_{\recsym}^1(\mathbf{v}) ) \notag \\
    & = T_{rot} J_{\Gamma_{T\recsym}^2}( T_{rot}^{-1} T_{rot} \Gamma_{\recsym}^1(\mathbf{v}) ) T_{rot}^{-1} \notag \\
    & = T_{rot} J_{\Gamma_{T\recsym}^2}( \Gamma_{\recsym}^1(\mathbf{v}) ) T_{rot}^{-1}
    \label{eq:JGamma2}
\end{align}
where in the second line we substituted Equation (\ref{eq:jac_simp}).  Taking determinants gives
\begin{align}
    \det \left( J_{\Gamma_{T\recsym}^2}(\Gamma_{T\recsym}^1(T\mathbf{v})) \right)
    & = \det \left( T_{rot} J_{\Gamma_{T\recsym}^2}( \Gamma_{\recsym}^1(\mathbf{v}) ) T_{rot}^{-1} \right) \notag \\
    & = \det \left( T_{rot}^{-1} T_{rot} J_{\Gamma_{T\recsym}^2}( \Gamma_{\recsym}^1(\mathbf{v}) )  \right) \notag \\
    & = \det \left( J_{\Gamma_{T\recsym}^2}( \Gamma_{\recsym}^1(\mathbf{v}) )  \right)
    \label{eq:jacobian_result_2}
\end{align}
where in the second line, we used the fact that permuting the order of a matrix multiplication does not affect the determinant.

Combining Equations (\ref{eq:distribution_result_1}), (\ref{eq:jacobian_result_1}), and (\ref{eq:jacobian_result_2}), we finally arrive at:
\begin{align}
    & p_{\mathbf{z}}( \Gamma_{T\recsym}^2(\Gamma_{T\recsym}^1(T\mathbf{v})) ) |\det J_{\Gamma_{T\recsym}^2}(\Gamma_{T\recsym}^1(T\mathbf{v})) | |\det J_{\Gamma_{T\recsym}^1}(T\mathbf{v}) |  \notag \\
    & \hspace{1.0cm} = \quad p_{\mathbf{z}}( \Gamma_{\recsym}^2(\Gamma_{\recsym}^1(\mathbf{v})) ) |\det J_{\Gamma_{\recsym}^2}(\Gamma_{\recsym}^1(\mathbf{v})) | |\det J_{\Gamma_{\recsym}^1}(\mathbf{v}) |
\end{align}
which is exactly Equation (\ref{eq:to_prove}).  Thus, we have shown that $p(TV | N, T\recsym) = p(V | N, \recsym)$, as desired.  

Let us turn now to the permutation case, which is quite similar.  Similar to Equation (\ref{eq:to_prove}), we need to show
\begin{align}
    & p_{\mathbf{z}}( \Gamma_{\hat{\pi}\recsym}^2(\Gamma_{\hat{\pi}\recsym}^1(\pi\mathbf{v})) ) |\det J_{\Gamma_{\hat{\pi}\recsym}^2}(\Gamma_{\hat{\pi}\recsym}^1(\pi\mathbf{v})) | |\det J_{\Gamma_{\hat{\pi}\recsym}^1}(\pi\mathbf{v}) |  \notag \\
    & \hspace{1.0cm} = \quad p_{\mathbf{z}}( \Gamma_{\recsym}^2(\Gamma_{\recsym}^1(\mathbf{v})) ) |\det J_{\Gamma_{\recsym}^2}(\Gamma_{\recsym}^1(\mathbf{v})) | |\det J_{\Gamma_{\recsym}^1}(\mathbf{v}) |
    \hspace{0.5cm} \text{for } \pi \in \mathbb{S}_n \text{ and } \hat{\pi} \in \mathbb{S}_{\hat{N}}
    \label{eq:perm_to_prove}
\end{align}
From Lemmata 1 and 2, we have that
\begin{equation}
    \Gamma_{\hat{\pi}\recsym}^1(\pi\mathbf{v}) = \pi \Gamma_{\recsym}^1(\mathbf{v}) \quad \text{and} \quad \Gamma_{\hat{\pi}\recsym}^2(\pi \mathbf{u}) = \pi \Gamma_{\recsym}^2(\mathbf{u})
\end{equation}
Thus
\begin{align}
    p_{\mathbf{z}}( \Gamma_{\hat{\pi}\recsym}^2(\Gamma_{\hat{\pi}\recsym}^1(\pi\mathbf{v})) )
    & = p_{\mathbf{z}}( \Gamma_{\hat{\pi}\recsym}^2(\pi \Gamma_{\recsym}^1(\mathbf{v})) ) \notag \\
    & = p_{\mathbf{z}}( \pi \Gamma_{\recsym}^2( \Gamma_{\recsym}^1(\mathbf{v})) ) \notag \\
    & = p_{\mathbf{z}}( \Gamma_{\recsym}^2( \Gamma_{\recsym}^1(\mathbf{v})) )
    \label{eq:perm_result_1}
\end{align}
where the last line follows from the permutation-invariance of the Gaussian distribution $ p_{\mathbf{z}}(\cdot)$.

In a manner parallel to the derivation of Equation (\ref{eq:JGamma1}), we can show that
\begin{equation}
    J_{\Gamma_{\hat{\pi}\recsym}^1}(\pi\mathbf{v}) = \boldsymbol{\pi} J_{\Gamma_{\recsym}^1}(\mathbf{v})
\end{equation}
where $\boldsymbol{\pi}$ now indicates the permutation matrix associated with the permutation $\pi$.  Thus, we have that
\begin{equation}
    \det\left( J_{\Gamma_{\hat{\pi}\recsym}^1}(\pi\mathbf{v}) \right) = \det(\boldsymbol{\pi}) \det\left( J_{\Gamma_{\recsym}^1}(\mathbf{v}) \right) = \pm \det\left( J_{\Gamma_{\recsym}^1}(\mathbf{v}) \right)
    \label{eq:perm_result_2}
\end{equation}
where we have used the fact that a permutation matrix has determinant of $\pm 1$.  Similarly, in a manner parallel to the derivation of Equation (\ref{eq:JGamma2}), we can show that 
\begin{equation}
    J_{\Gamma_{\hat{\pi}\recsym}^2}(\Gamma_{\hat{\pi}\recsym}^1(\pi\mathbf{v})) = \boldsymbol{\pi} J_{\Gamma_{\recsym}^2}(\Gamma_{\recsym}^1(\mathbf{v})) \boldsymbol{\pi}^{-1}
\end{equation}
so that
\begin{align}
    \det\left( J_{\Gamma_{\hat{\pi}\recsym}^2}(\Gamma_{\hat{\pi}\recsym}^1(\pi\mathbf{v})) \right)
    & = \det\left( \boldsymbol{\pi} J_{\Gamma_{\recsym}^2}(\Gamma_{\recsym}^1(\mathbf{v})) \boldsymbol{\pi}^{-1} \right) \notag \\
    & = \det\left( \boldsymbol{\pi}^{-1} \boldsymbol{\pi} J_{\Gamma_{\recsym}^2}(\Gamma_{\recsym}^1(\mathbf{v})) \right) \notag \\
    & = \det\left( J_{\Gamma_{\recsym}^2}(\Gamma_{\recsym}^1(\mathbf{v})) \right)
    \label{eq:perm_result_3}
\end{align}
Combining Equations (\ref{eq:perm_result_1}), (\ref{eq:perm_result_2}), and (\ref{eq:perm_result_3}) yields Equation (\ref{eq:perm_to_prove}); completing the proof.  \qedsymbol

\section{The Edge and Properties Distributions}
\label{sec:edges_and_properties}
\parbasic{The Edge Distribution} Given the invariance conditions described in Equation (\ref{eq:invariance}), we propose a distribution which displays \textit{conditional independence}: 
$p\left( E = ( \mathbf{e}_{ij} \right)_{i<j: j \in \eta_i} | N, V, \recsym ) = 
    \prod_{i<j: j \in \eta_i} p( \mathbf{e}_{ij} | N, V, \recsym )$.
We opt for conditional independence for two reasons: (1) The usual Markov decomposition of the probability distribution with terms of the form $p( \mathbf{e}_{ij} | \mathbf{e}_{<ij}, N, V, \recsym )$ implies a particular ordering of the edges, and is therefore not permutation-invariant. (2) $V$ is a deterministic and invertible function of the flow's noise vector $\mathbf{z}$; thus, conditioning on $V$ is the same as conditioning on $\mathbf{z}$.  If $E$ is a deterministic (but not necessarily invertible) function of $\mathbf{z}$, then conditional independence is correct.

To compute $p( \mathbf{e}_{ij} | N, V, \recsym )$, we use a second \conditionalEGNN{}.  The key distinction between this network and the \conditionalEGNN{} used in computing the vertex distribution is the initial conditions.  In the case of the vertex distribution, the initial conditions are $\mathbf{x}_i^1 = \mathbf{0}$ and $\mathbf{h}_i^1 = \mathbf{0}$.  In the current case of the edge distribution, we are given $V$ (we are conditioning on it); thus, we take the initial conditions to be $\bt{x}_i^1 = \mathbf{x}_i(V)$ and $\bt{h}_i^1 = \mathbf{h}_i(V)$.  In other words, the initial values are given the vertex list $V$ itself.

Given this second \conditionalEGNN{}, we can compute the edge distribution as
\begin{equation}
    p\left( \mathbf{e}_{ij} | N, V, \recsym \right) = \mathbf{e}_{ij}^T \, \texttt{MLP}\left(\bt{m}^L_{ij}\right)
\end{equation}
in the case of categorical properties (where $\texttt{MLP}$'s output is a softmax with $d_e$ entries); analogous expressions exist for ordinal or continuous properties.  It is straightforward to see that this distribution satisfies the invariance properties in Equation (\ref{eq:invariance}).  The corresponding loss function is a simple cross-entropy loss (or regression loss for non-categorical properties).

\parbasic{The Property Distribution} We propose the following distribution.  We use a standard Markov decomposition:
$
    p(A | N, V, E, \recsym) = \prod_{k=1}^K p\left( \mathbf{a}_k | \mathbf{a}_{1:(k-1)}, N, V, E, \recsym \right)
$.
Let
\begin{equation}
    \boldsymbol{\xi}_h = \frac{1}{N} \sum_{i=1}^N \texttt{MLP} \left( \bt{h}^L_i \right)
    \hspace{0.8cm}
    \boldsymbol{\xi}_e = \frac{1}{|E|} \sum_{i<j: j \in \eta_i} \texttt{MLP} \left( \mathbf{e}_{ij} \right)
    \hspace{0.8cm}
    \boldsymbol{\xi}_{a,k} = \texttt{MLP}\left( \sum_{j=1}^{k-1} W_j \mathbf{a}_j \right)
\end{equation}
where the matrices $W_1, \dots W_K$ all have the same number of rows.
Then we set
\begin{equation}
    p\left( \mathbf{a}_k \left| \mathbf{a}_{1:(k-1)}, N, V, E, \recsym \right. \right) = \mathbf{a}_k^T \texttt{MLP} \left( \texttt{concat}\left( \boldsymbol{\xi}_h , \boldsymbol{\xi}_e, \boldsymbol{\xi}_{a,k} \right) \right)
\end{equation}
in the case of categorical properties; analogous expressions exist for ordinal or continuous properties.  Note that the only item which changes for the different properties $k$ is the vector $\boldsymbol{\xi}_{a,k}$. It is easy to see that this distribution satisfies the invariance properties in Equation (\ref{eq:invariance}).  The corresponding loss function is a simple cross-entropy loss (or regression loss for non-categorical properties).

\section{Experimental Setup}
\label{sec:experimental_setup}
\parbasic{Molecular Dataset} Datasets with a large number of receptor-ligand complexes are critical to our endeavour.  Many models have relied on the high quality PDBbind dataset which curates the Protein Data Bank (PDB) \citep{liu2017forging}; however, for the training of generative models, this dataset is relatively small.  CrossDocked2020 \citep{francoeur2020three} is the first large-scale standardized dataset for training ML models with ligand poses cross-docked against non-cognate receptor structure, greatly expanding the number of poses available for training. The dataset is organized by clustering of similar binding pockets across the PDB; each cluster contains ligands cross-docked against all receptors in the pocket. Each receptor-ligand structure also contains information indicating the nature of the docked pair, such as root mean squared deviation (RMSD) to the reference crystal pose and Vina cross-docking score \citep{trott2010autodock} as implemented in Smina \citep{koes2013lessons}. The dataset contains 22.5 million poses of ligands docked into multiple similar binding pockets across the PDB.  We use the authors' suggested split into training and validation sets \citep{francoeur2020three}.  This dataset contains docked receptor-ligand pairs whose binding pose RMSD is lower than $2\text{\AA}$.  Based on considerations of training duration, we keep only those data points whose ligand has 30 atoms or fewer with atom types in \{C, N, O, F\}. We also keep only data points with potentially valid ligand-protein binding properties, i.e. whose ligands do not contain duplicate vertices and whose predicted Vina scores \citep{trott2010autodock} are within distribution.  The refined datasets consist of 132,863 training data points and 63,929 validation data points.

\parbasic{Features} The ligand features that we wish to predict include the atom type $\in$ \{C, N, O, F\} (categorical); the stereo parity $\in$ \{not stereo, odd, even\} (categorical); and charge $\in \{-1, 0, +1\}$ (ordinal).  The receptor features that are used are computed with the Graphein library \citep{jamasb2022graphein}.  The vertex features include: the atom type $\in$ \{C, N, O, S, ``other''\} where ``other'' is a catch-all for less common atom types\footnote{Specifically: Na, Mg, P, Cl, K, Ca, Co, Cu, Zn, Se, Cd, I, Hg.} (categorical); the Meiler Embeddings \citep{meiler2001generation} (continuous $\in \mathbb{R}^{7}$).  The bond (edge) properties include: the bond order $\in$ \{Single, Double, Triple\} (categorical); covalent bond length (continuous $\in \mathbb{R}$). The receptor overall graph properties ($\hat{A}$) contain the weight of all chains contained within a polypeptide structure, see \citep{jamasb2022graphein}.

\parbasic{Training} Training the model takes approximately 18 days using a single NVIDIA A100 GPU for 39 epochs. We proactively stopped the training procedure when the negative log-likelihood (NLL) term reached a low improvement rate of between epochs. We train with the Adam optimizer, weight decay of $10^{-12}$, batch size of 128, and learning rate of $2\times10^{-4}$. To improve the stability of the continuous flow model and to deal with ODE stiffness issues we use the ODE regularization term described in \citep{finlay2020train} with a value of $10^{-3}$; and perform gradient clipping, where the clipping term is set by calculating a moving average of 50 steps of the normalizing flows gradient-norm. This is parallel to the treatment in \citep{satorras2021flows}. We use dopri5 (Runge-Kutta 4(5) of Dormand-Prince) as the ODE solver with relative (rtol) and absolute (atol) error tolerance values of $10^{-4}$.

\section{Detailed Experimental Results}
\label{sec:detailed_experimental_results}
We first evaluate the utility of our technique as a conditional generative model. Given a receptor, we generate ligands which may successfully bind to that receptor. To show that our method is competitive as a generative model we compare against \citet{liu2022generating} -- one of the recent works that showed promising results in similar settings. We then demonstrate the utility of our technique for indicating the nature of the interaction of ligands with their corresponding binding site. The trained model provides the conditional probability, $p(\ligsym | \recsym)$, which defines the likelihood of a binding event occurring given a ligand, based on the presence of a specific target receptor.

\parbasic{Conditional Generative Model} We perform inference using the method suggested in the baseline \citep{liu2022generating}.  Given a receptor, we sample from the learned distribution, which generates the ligands' vertices; we then then apply OpenBabel \citep{hummell2021novel} to construct bonds.  Evaluation follows the standard procedure \citep{ragoza2022generating,liu2022generating}.  First, the receptor target is computed by taking all of the atoms in the receptor that are less than $15\text{\AA}$ from the center of mass of the reference ligand.  We then generate 100 ligands for each reference binding site in the evaluation set, and compute statistics (i.e. validity and $\Delta$Binding, see below) on this set of samples. As in \citep{ragoza2022generating,liu2022generating}, 10 target receptors for evaluation; each target receptor has multiple associated ligands, leading to 90 (receptor, reference-ligand) pairs. We train the baseline technique \citep{liu2022generating} on our filtered dataset. More specifically, the training continues for 100 epochs, using the hyperparameters given in the paper, with one exception: we set the atom number range of the autoregressive generative process according to the atom distribution of the filtered dataset.

\begin{figure}[tb]

\centering
\includegraphics[width=.25\textwidth]{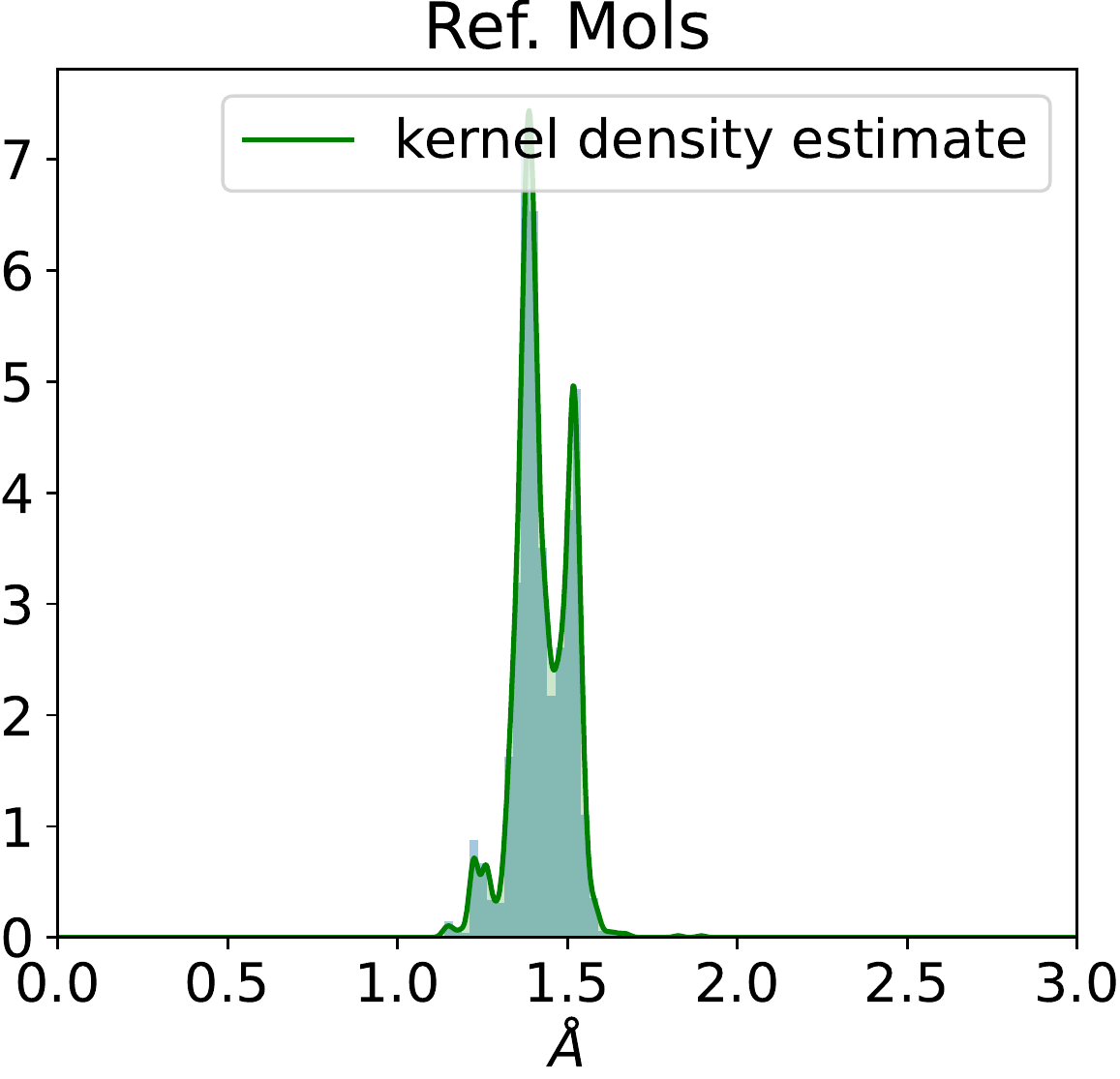}\hfill
\includegraphics[width=.25\textwidth]{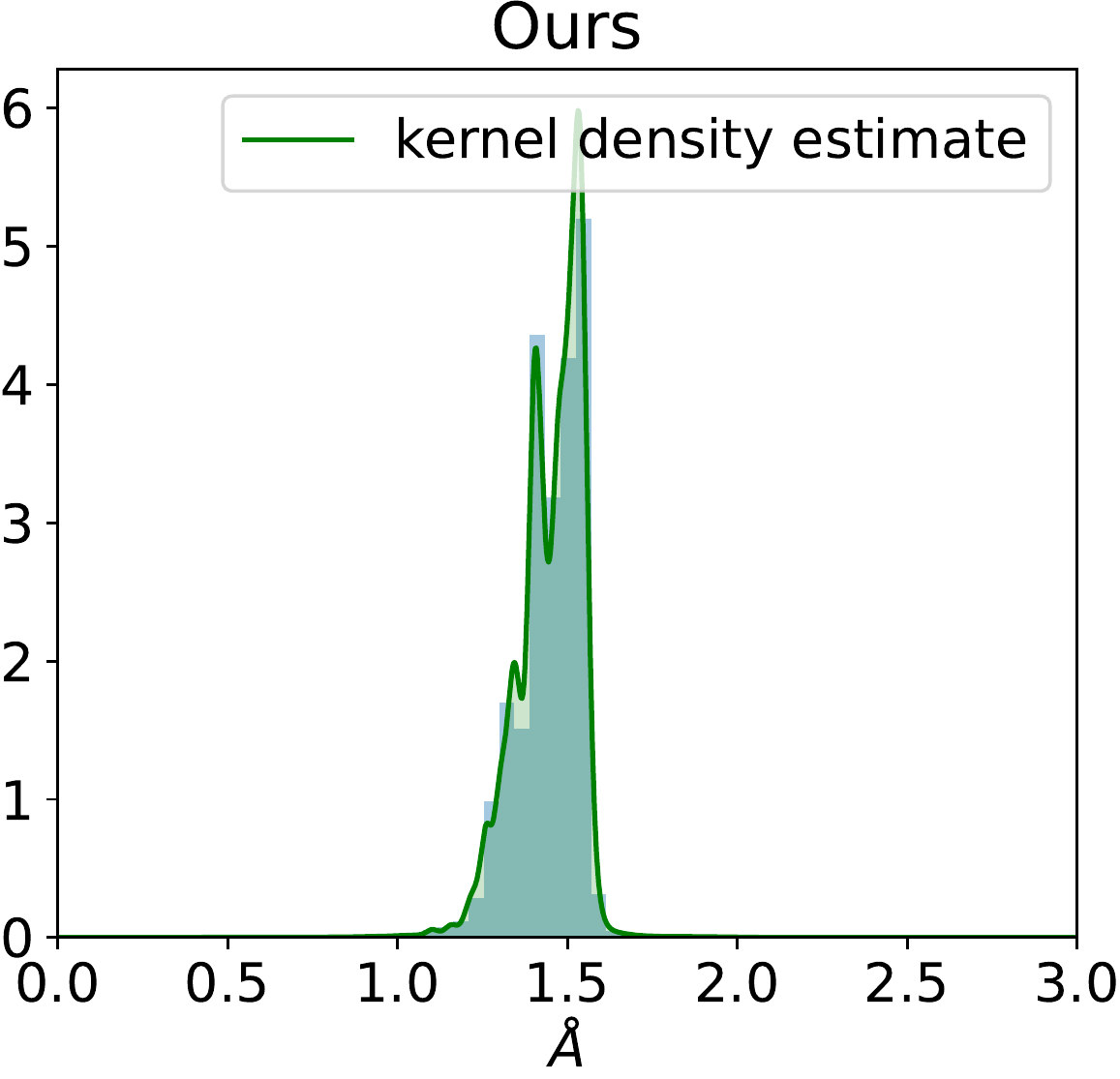}\hfill
\includegraphics[width=.25\textwidth]{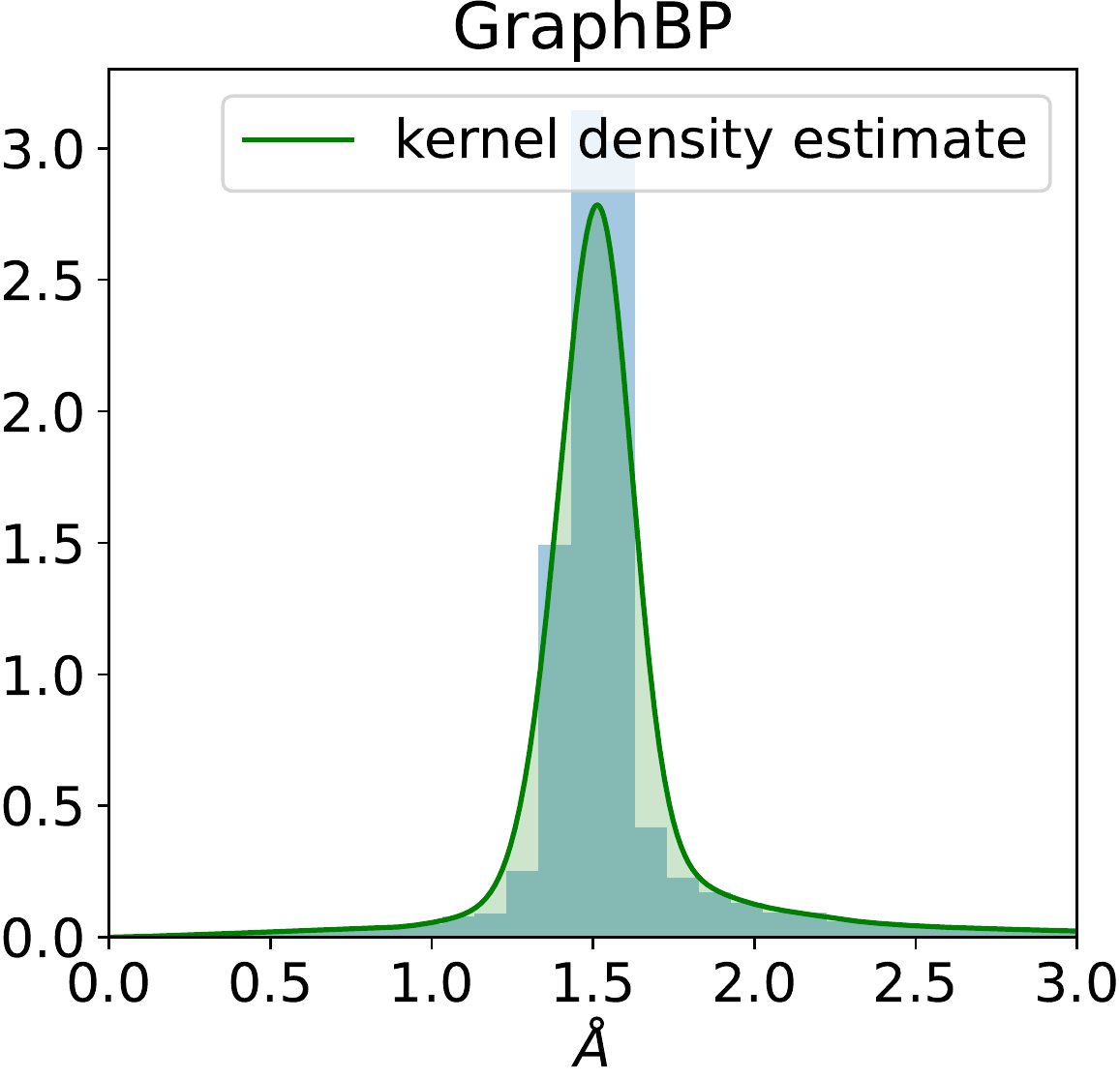}

\caption{Normalized histogram of relative distances between atoms}
\label{fig:bond_dist}

\end{figure}

\textbf{\textit{Validity}} The validity is defined as the percentage of molecules that are chemically valid among all generated molecules. A molecule is valid if it can be sanitized by RDKit; for an explanation of the sanitization procedure, see \citep{Landrum2016RDKit2016_09_4}. As shown in Table \ref{tab:validity_binding_full}(a), our model produces ligands with a validity of 99.87\%, surpassing the baseline, GraphBP. We also compute the distribution of bond distances of the two methods, and compare this to distribution of the reference ligands; see Figure \ref{fig:bond_dist}.  Our method's distribution is considerably closer to the reference distribution than GraphBP; some non-trivial fraction of the time, GraphBP produces unusual, very high bond distances.  (In fact, we have discarded values higher than $10\text{\AA}$ on the GraphBP plot so as to display the distributions on similar scales.)  This impression is reinforced in Table \ref{tab:validity_binding_full}(a) which compares the mean and standard deviation of these distributions.

\begin{table}[tb]
    \begin{subtable}{.5\linewidth}
      \centering
        \begin{tabular}{ |c||c|c|c|} 
            \hline
            \multicolumn{4}{|c|}{\textbf{Validity}} \\
            \hline
            \multicolumn{2}{|c|}{Ours} & \multicolumn{2}{|c|}{GraphBP} \\
            \hline
            \multicolumn{2}{|c|}{99.87\%} & \multicolumn{2}{|c|}{99.75\%} \\
            \hline
            \hline
            \hline
            \multicolumn{4}{|c|}{\textbf{Bond Length Distribution}} \\
            \hline
            & Ref. Mols. & Ours & GraphBP \\
            \hline
            mean & 1.42 & 1.45 & 1.65 \\
            std & 0.08 & 0.10 & 0.95 \\
            \hline
        \end{tabular}
        \caption{Ligand validity and bond length distribution}
    \end{subtable}%
    \begin{subtable}{.5\linewidth}
      \centering
        \begin{tabular}{ |c||c|c|c|} 
            \hline
            \multicolumn{4}{|c|}{$\mathbf{\Delta}$\textbf{Binding}} \\
            \hline
            \multicolumn{2}{|c|}{Ours} & \multicolumn{2}{|c|}{GraphBP} \\
            \hline
            \multicolumn{2}{|c|}{35.7\%} & \multicolumn{2}{|c|}{22.76\%} \\
            \hline\hline\hline
            \multicolumn{4}{|c|}{\textbf{Predicted Affinity Distribution}} \\
            \hline
            & Ref. Mols. & Ours & GraphBP \\
            \hline
            mean & 5.09 & 4.56 & 4.31 \\
            std & 1.16 & 1.05 & 1.03 \\
            \hline
        \end{tabular}
        \caption{$\Delta$Binding and predicted affinity distribution}
    \end{subtable}
    \caption{Comparison of molecule validity and $\Delta$Binding between proposed method and GraphBP.}
    \label{tab:validity_binding_full}
\end{table}

\textbf{\textit{Binding Affinity}} A more interesting measure than validity is $\Delta$Binding, which measures the percentage of generated molecules that have higher predicted binding affinity to the target binding site than the corresponding reference molecule.  To compute binding affinities, we follow the procedure used by GraphBP.  Briefly, we refine the generated 3D molecules by Universal Force Field (UFF) minimization \citep{rappe1992uff}; then, Vina minimization and CNN scoring are applied to both generated and reference molecules by using gnina, a molecular docking program \citep{mcnutt2021gnina}.  As can be seen in Table \ref{tab:validity_binding_full}(b), our result improves significantly on the baseline.  Raw GraphBP attains $\Delta$Binding = 13.45\%.  By playing with the minimum and the maximum atom number of the baseline autoregressive model, we were able to improve this to 22.76\%; however, note that this results in a  reduction in validity from 99.75\% to 99.54\%.  Our method attains $\Delta$Binding = 35.7\%, which is a relative improvement of 56.81\% over the better of the two GraphBP scores.

\begin{figure*}
    \centering
    \begin{tabular}{cccccc}
         \fig{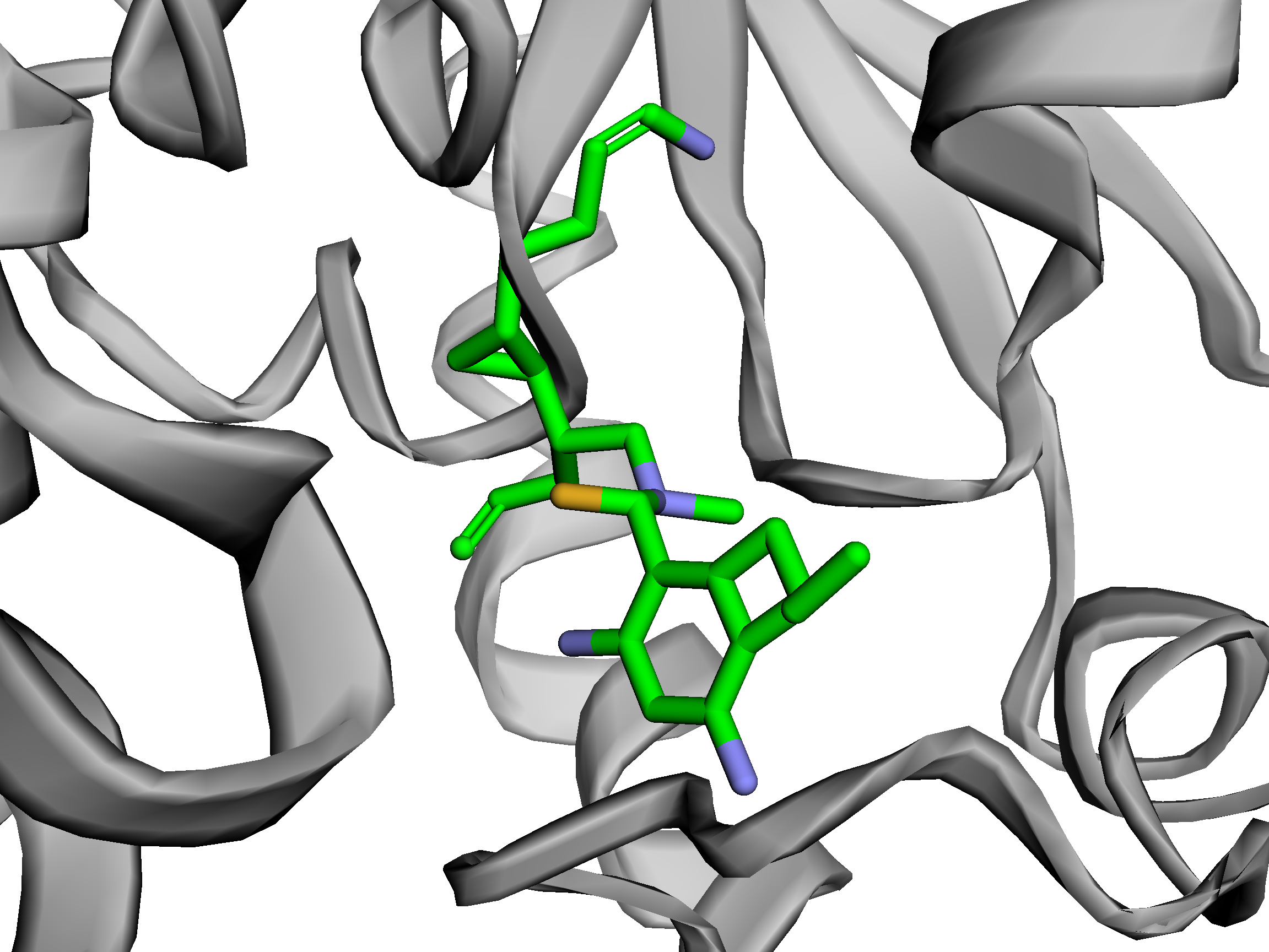}{0.13}
         & \fig{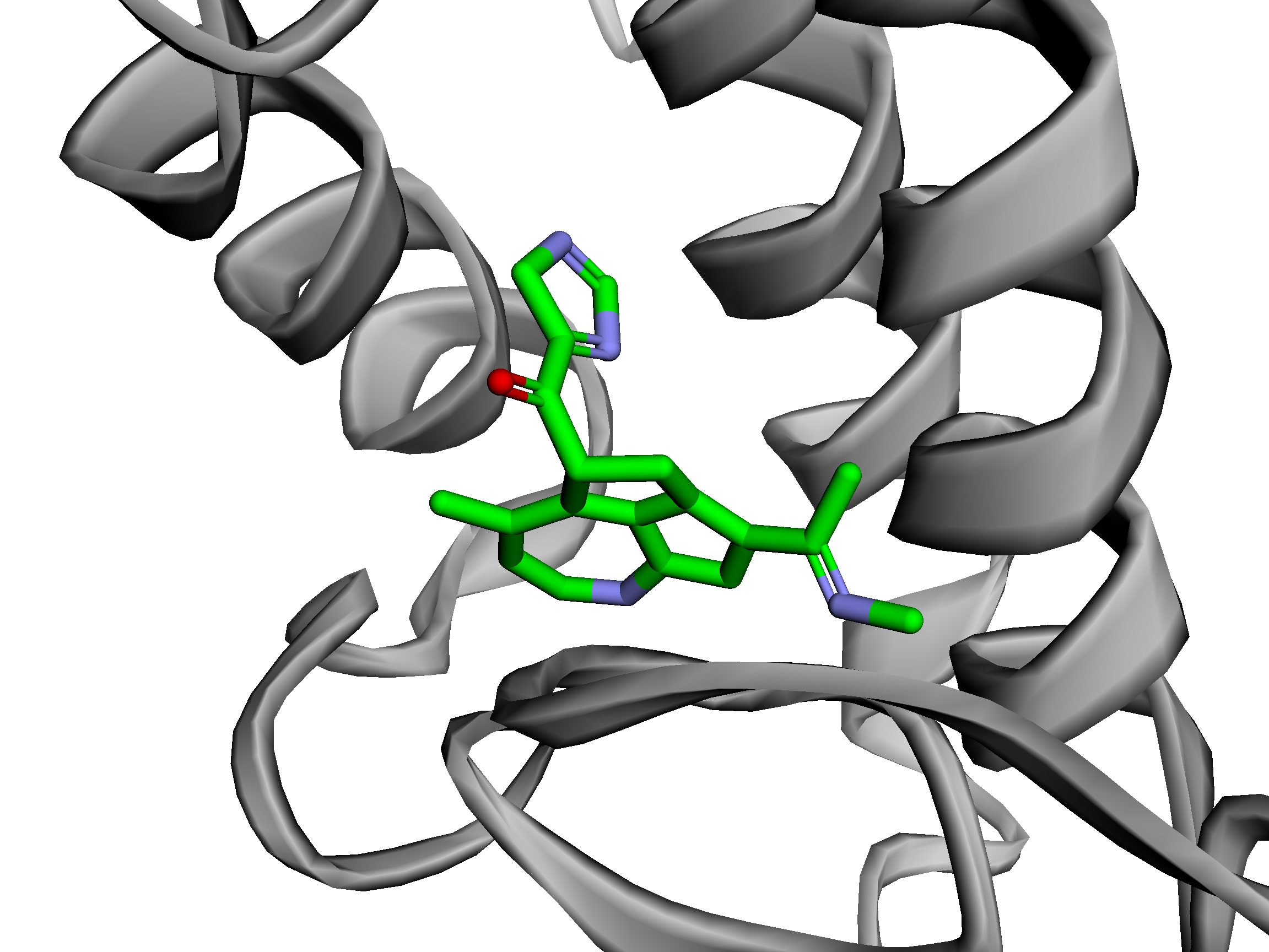}{0.13}
         & \fig{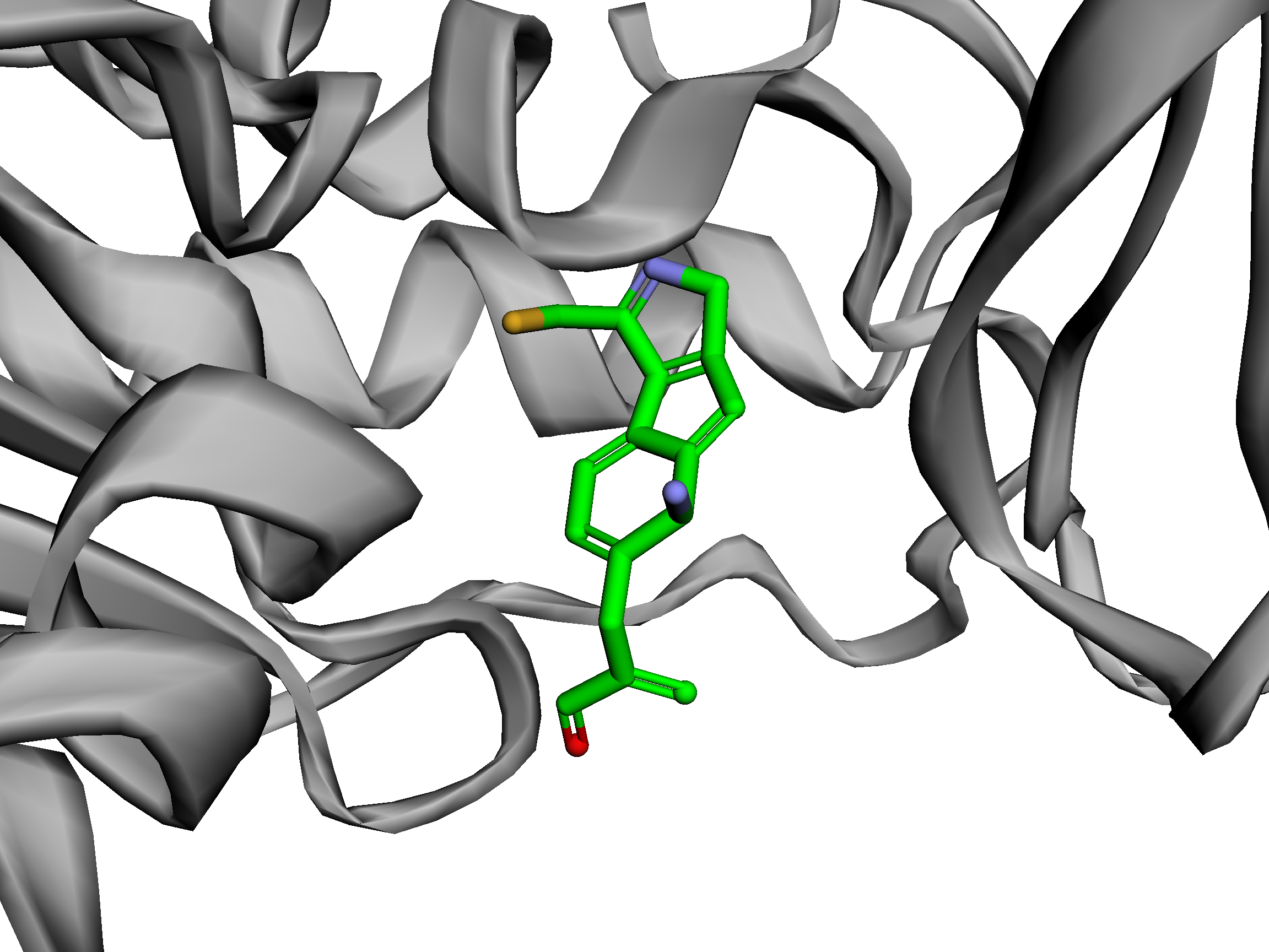}{0.13}
         & \fig{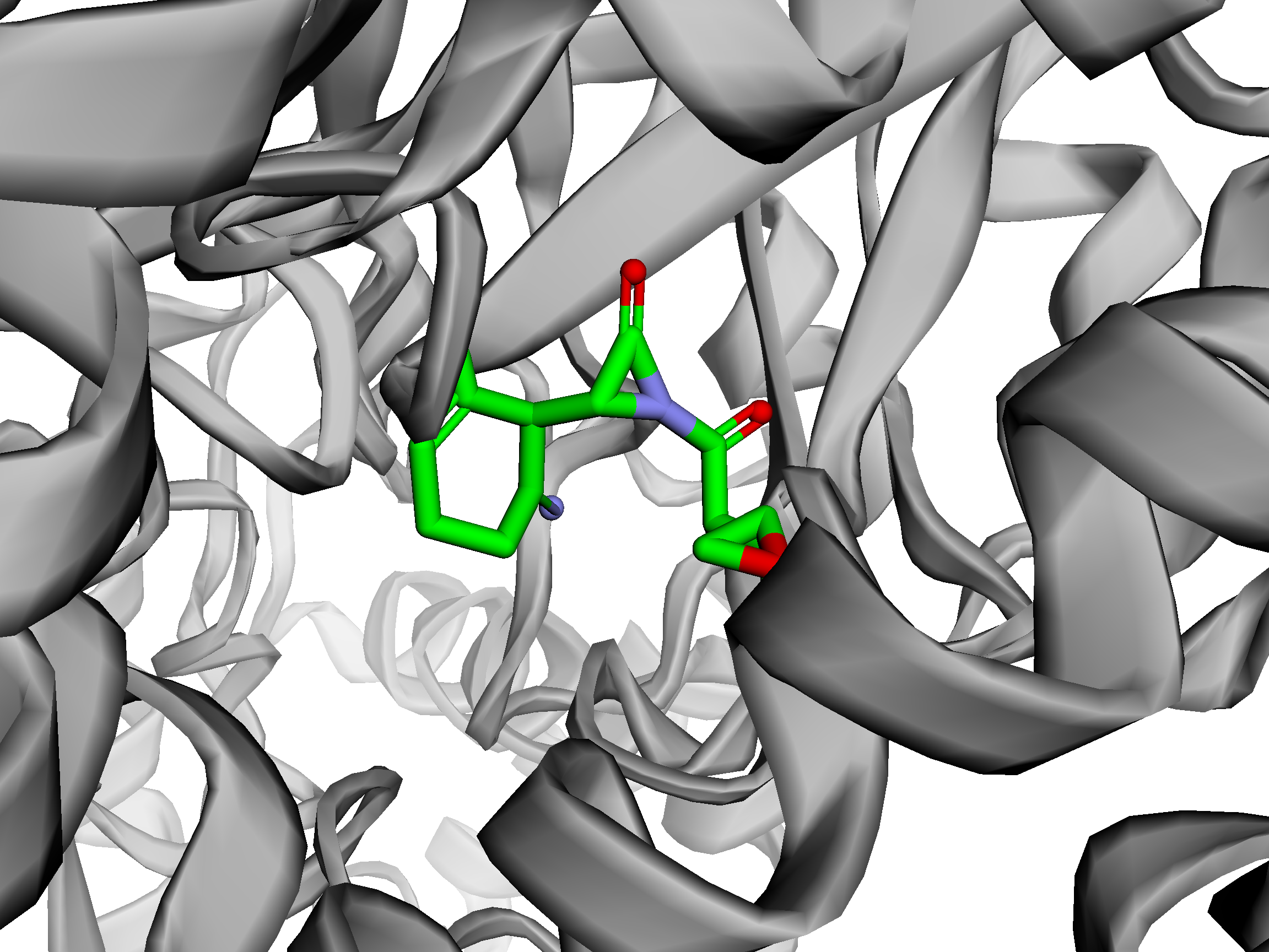}{0.13}
         & \fig{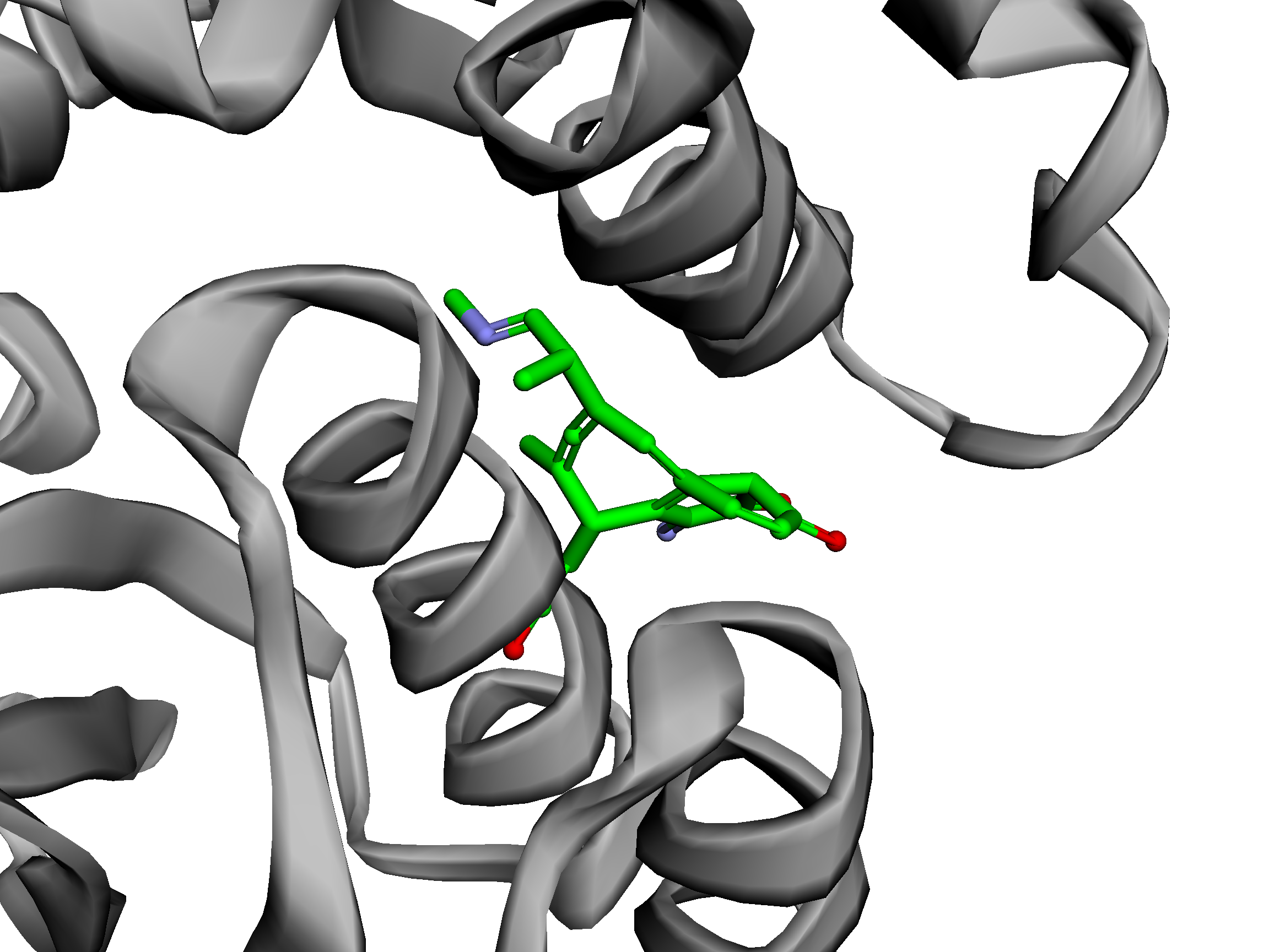}{0.13}
         & \fig{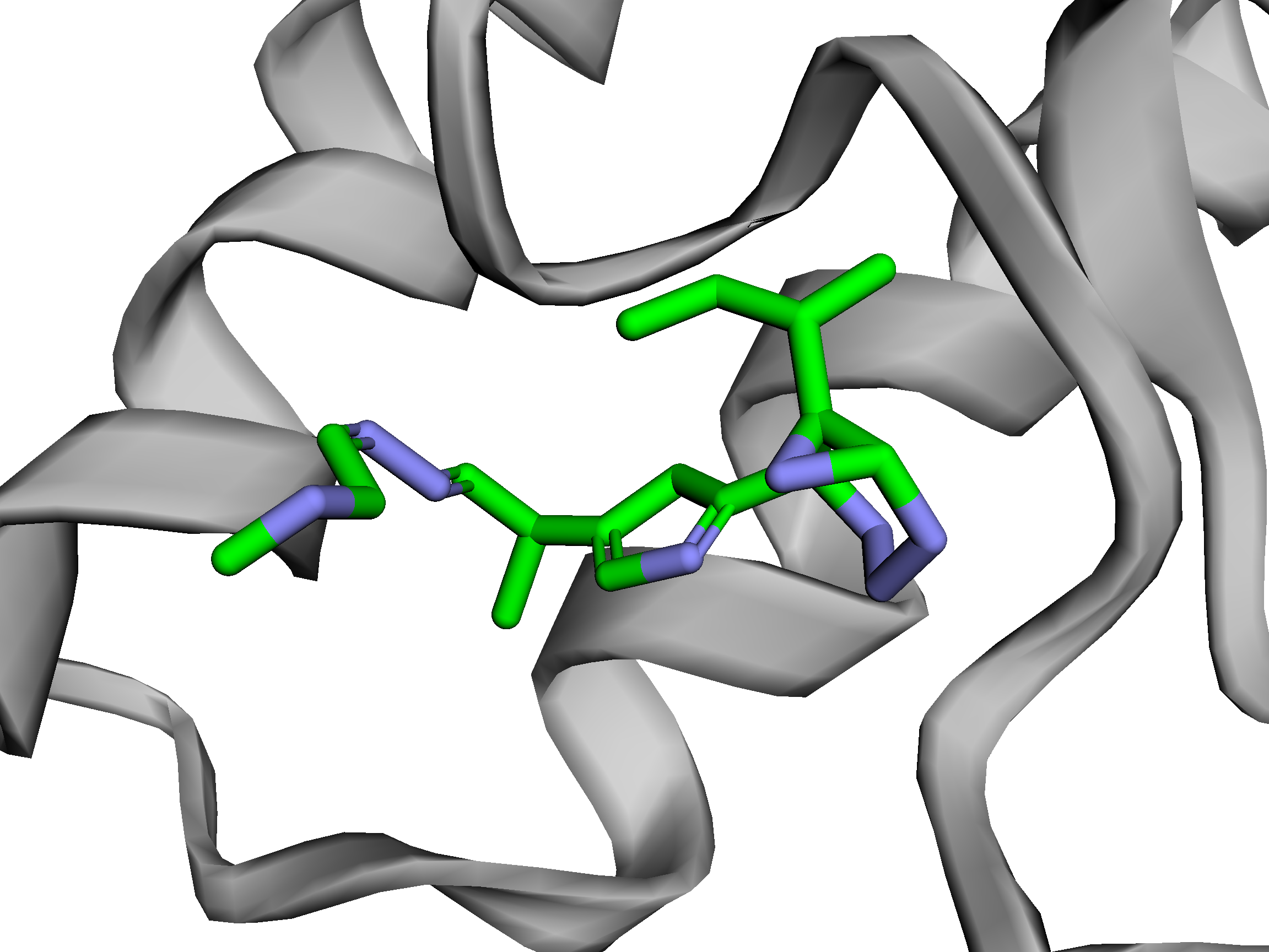}{0.13}
         \\
         \fig{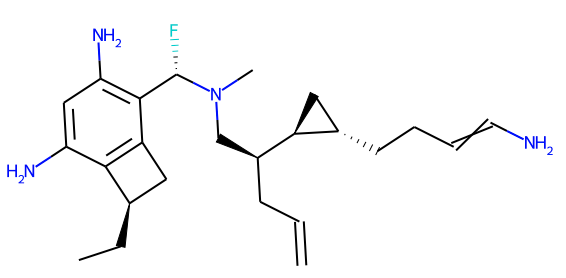}{0.13}
         & \fig{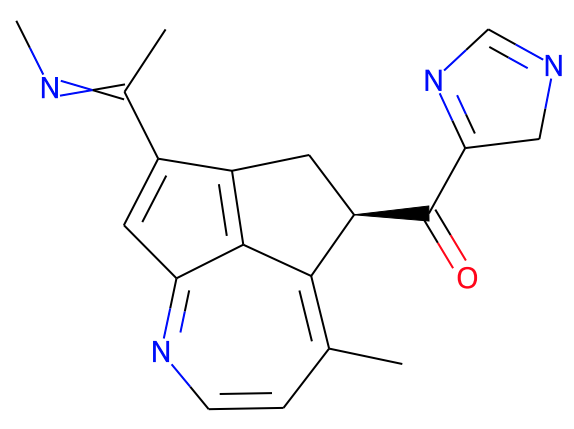}{0.13}
         & \fig{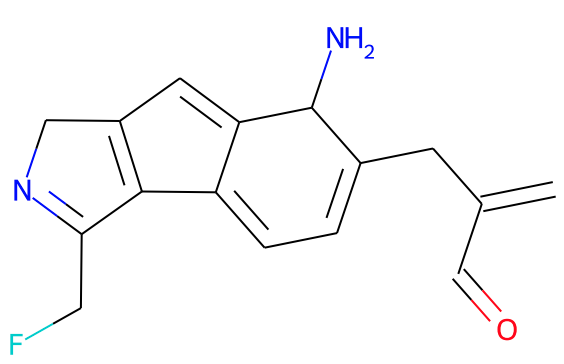}{0.13}
         & \fig{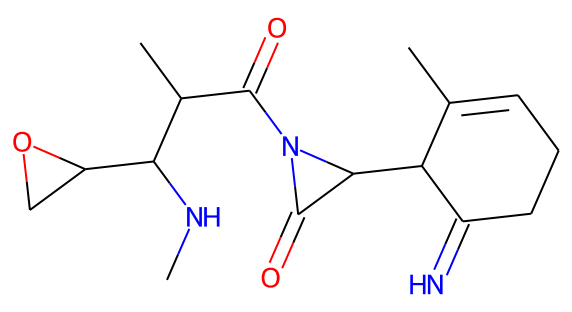}{0.13}
         & \fig{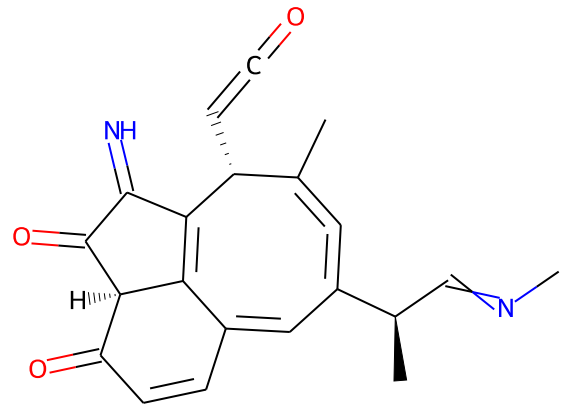}{0.13}
         & \fig{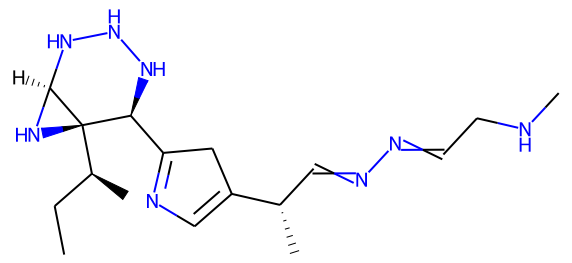}{0.13}
         \\
         \fig{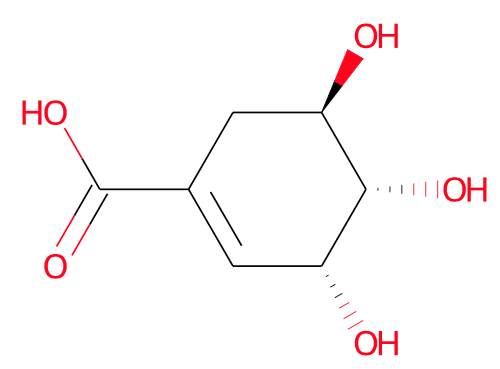}{0.13}
         & \fig{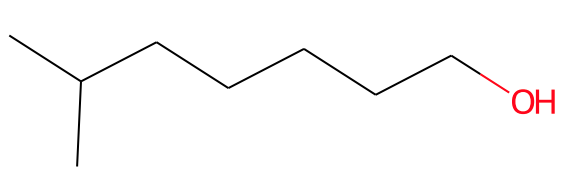}{0.13}
         & \fig{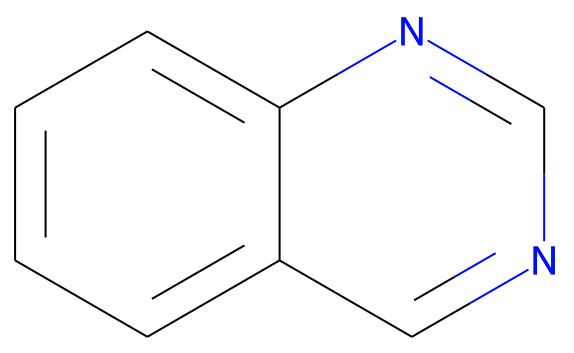}{0.13}
         & \fig{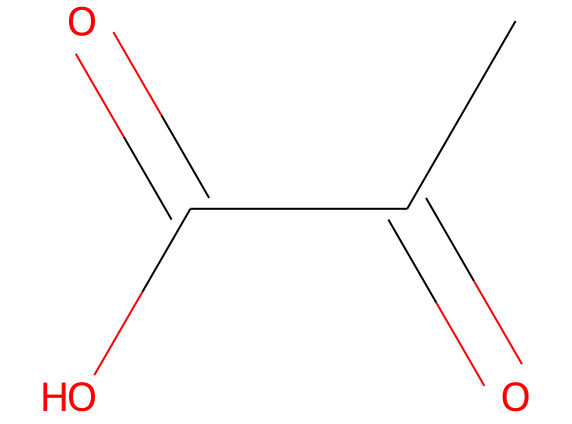}{0.13}
         & \fig{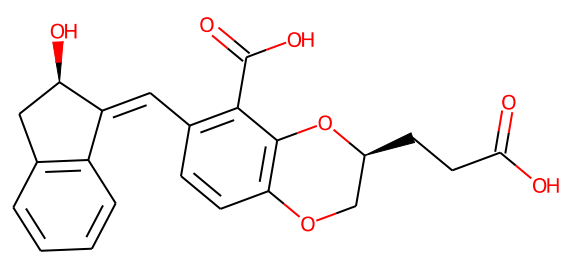}{0.13}
         & \fig{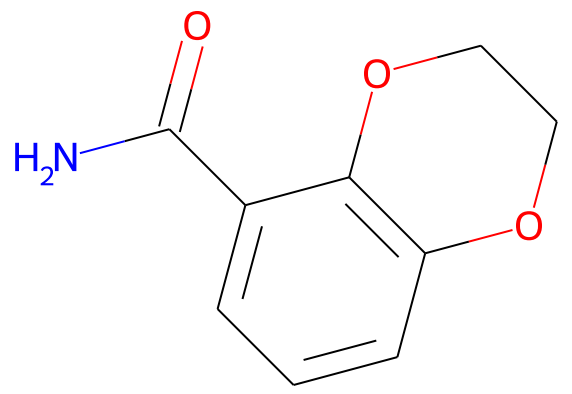}{0.13}
    \end{tabular}
    \caption{Comparison between generated 3D molecules for target binding-site and reference molecules. Receptor IDs, left to right: 1zyu, 2qu9, 2hw1, 3lcg, 3zt3, 5lvq.  Top: generated ligand (colour) + receptor. Middle: generated ligand chemical structure.  Bottom: reference ligand chemical structure.}
    \label{fig:qualitative}
\end{figure*}

\textbf{\textit{Qualitative Results}} We show examples of generated ligands in Figure \ref{fig:qualitative}, along with their chemical structures.  Note that the structures of the generated molecules differ substantially from the reference molecules, indicating that the model has indeed learn to generalize to interesting novel structures.

\parbasic{Binding Likelihood} The CrossDocked2020 dataset \citep{francoeur2020three} includes quantitative measures indicating the quality of the binding of each docked receptor-ligand structure: (i) root mean squared deviation (RMSD) to the reference crystal pose; (ii) Vina cross-docking score \citep{trott2010autodock} as implemented in Smina \citep{koes2013lessons}. Scoring functions that represent and predict ligand-protein interactions are important for applications in structure-based drug discovery (e.g. energy minimization, molecular dynamics simulations, and hit identification/lead optimization \citep{dewitte1996smog, mcinnes2007virtual,charifson1999consensus}), and particularly important for molecular docking where one seeks to predict the probability that binding occurs given a specified orientation and conformation (i.e. pose) of a ligand with respect to a target receptor \citep{su2018comparative, wang2003comparative, kitchen2004docking,warren2006critical,cheng2009comparative,huang2011scoring,trott2010autodock, cheng2012structure}. We show how the conditional probability of our model may also be used as such indicator.
As we have seen in Section \ref{sec:methods}, our method is not trained on either of the binding quality indicators (Vina / RMSD), but instead is the result of learning a conditional probability distribution from ligand-receptor pairs; therefore it may be viewed as a complementary scoring method. Specifically, given the conditional probability $p(\ligsym | \recsym)$ from the trained model, its negative log-likelihood (NLL) may be used a scoring method.  Such an approach can be considered as a member of the family of knowledge-based methods \citep{muegge2000knowledge, gohlke2000knowledge,durrant2011nnscore, huang2011scoring, ballester2010machine,hassan2018dlscore,wojcikowski2019development} which are constructed from entirely non-physical statistical potentials derived from known receptor-ligand complexes.

\textbf{\textit{Vina and RMSD}} We split the 63,929 validation data points into three different groups according their Vina scores: (a) Vina $\in [0, 5]$ (b) Vina $\in (5, 9]$ (c) Vina $\in (9, 15]$.  We randomly sample 10,000 receptor-ligand complexes from each group.  Using the trained model, we calculate the average NLL for each group separately, where lower NLL indicates better conditional likelihood. The results in Table \ref{tab:conditional_probability}(a) show good correspondence between the Vina scores (higher values describe better affinity) with the binding likelihood outcomes of our model (lower NLL is better).

We then repeat this experiment, but using RMSD values instead of Vina scores.  Again, we divide the validation set into three groups according to their RMSD values: (a) RMSD $\in [0, 1]$ (b) RMSD $\in (1, 1.5]$ (c) RMSD $\in (1, 2]$ (all values in Angstroms).  Again, we see a good correspondence between the RMSD values (lower values describe a better connected complex) with the binding likelihood outcomes of our model (lower NLL is better).

\textbf{\textit{Rigid Transformations}} The nature of binding between receptor and corresponding ligand is affected by the position of the ligand in the binding site. The ligand pose for each ligand-receptor pair in the validation data points was refined using the UFF force-field and then optimized with respect to the receptor structure using the Vina scoring function. We randomly select 10,000 validation data points and apply a rigid body transformation solely to the minimized ligand-graph pose. Any such changes to the ligand pose with respect to the binding site potentially affect and reduce the binding affinity. The results in Table \ref{tab:conditional_probability}(b) show the NLL under varying degrees of rigid transformations. The more significant the transformation, the lower the binding likelihood (i.e. the higher the NLL) gets. 

\begin{table}[tb]
    \begin{subtable}{1\linewidth}
      \centering
        \begin{tabular}{ |c||c|c|c|} 
            \hline
            \multicolumn{4}{|c|}{\textbf{Vina Score} $\uparrow$} \\
            \hline
            & \scriptsize{0 < Vina < 5} & \scriptsize{5 < Vina < 9} & \scriptsize{9 < Vina < 15} \\
            \hline
            NLL $\downarrow$ & \scriptsize{-77.83} & \scriptsize{-144.30} & \scriptsize{-180.37} \\
            \hline\hline\hline
            \multicolumn{4}{|c|}{\textbf{RMSD} $\downarrow$} \\
            \hline
            & \scriptsize{0 < RMSD < 1} & \scriptsize{1 < RMSD < 1.5} & \scriptsize{1.5 < RMSD < 2} \\
            \hline
            NLL $\downarrow$ & \scriptsize{-137.70} & \scriptsize{-135.72} & \scriptsize{-130.67} \\
            \hline
        \end{tabular}
        \caption{Ligand-receptor Vina score and RMSD vs average negative log-likelihood (NLL).}
    \end{subtable}
    
    \bigskip
    \begin{subtable}{1\linewidth}
      \centering
      \begin{tabular}{|c c|c|c|c|c|c|c|}
        \hline
        \multicolumn{8}{|c|}{\textbf{NLL under rigid transformation} $\downarrow$} \\
        \hline\hline
        \multicolumn{3}{|c|}{} &\multicolumn{5}{|c|}{Rotation} \\
        \multicolumn{3}{|c|}{} &\multicolumn{5}{|c|}{\tiny{$\theta_{x}\sim U(-\frac{\pi}{u}, \frac{\pi}{u}), \theta_{y}\sim U(-\frac{\pi}{2u}, \frac{\pi}{2u}), \theta_{z}\sim U(-\frac{\pi}{u}, \frac{\pi}{u})$} } \\ \cline{4-8}
        \multicolumn{3}{|c|}{}  & \scriptsize{} & \scriptsize{$u=20$} & \scriptsize{$u=10$} & \scriptsize{$u=\frac{20}{3}$} & \scriptsize{$u=5$}\\\hline
        \multirow{5}{*}{\rotatebox{90}{Translation}} & \multirow{5}{*}{\rotatebox{90}{\tiny{$\tau \sim U\left([-t, t]^3\right)$}}}  & \scriptsize{} & \scriptsize{\(-134.34\)} &  \scriptsize{\(-132.18 \)}&  \scriptsize{\(-124.15\)} &  \scriptsize{\(-110.45\)} &  \scriptsize{\(-91.94\)} \\\cline{3-8}
        & & \scriptsize{$t=2\text{\AA}$} & \scriptsize{\(-133.98\)} & \scriptsize{\(-130.80\)} & \scriptsize{\(-122.49\)} & \scriptsize{\(-109.22\)} &  \scriptsize{\(-90.75\)} \\ \cline{3-8}
        & & \scriptsize{$t=5\text{\AA}$} & \scriptsize{\(-129.28\)} & \scriptsize{\(-126.28\)} & \scriptsize{\(-118.57\)} & \scriptsize{\(-104.94\)} &  \scriptsize{\(-86.02\)} \\ \cline{3-8}
        & & \scriptsize{$t=10\text{\AA}$} & \scriptsize{\(-113.99\)} & \scriptsize{\(-111.80\)} & \scriptsize{\(-103.74\)} & \scriptsize{\(-91.15\)} &   \scriptsize{\(-71.83\)} \\ \cline{3-8}
        & & \scriptsize{$t=15\text{\AA}$} & \scriptsize{\(-88.85\)} & \scriptsize{\(-87.26\)} & \scriptsize{\(-76.85\)} &   \scriptsize{\(-66.19\)} &   \scriptsize{\(-45.89\)} \\ 
        \hline  
      \end{tabular}
        
        \caption{Average negative log-likelihood (NLL) under varying degrees of 3D rigid body transformations (translation and rotation) of ligand pose. $\theta_{x}, \theta_{y}, \theta_{z}$ are rotational Euler angles and $\tau \in \mathbb{R}^{3}$ is translation factor.}
    \end{subtable}
    \caption{Binding Likelihood. $\downarrow$ ($\uparrow$) indicates stronger binding for lower (higher) values.}
    \label{tab:conditional_probability}
\end{table}

\end{document}